# Title: Advancing operational PM2.5 forecasting with dual deep neural networks (D-DNet)

**Authors:** Shengjuan Cai[1,2], Fangxin Fang[1,2]*, Vincent-Henri Peuch[3], Mihai Alexe[3], Ionel Michael Navon[4], Yanghua Wang[1,2]

**Affiliations:**

[1]Resource Geophysics Academy, Imperial College London; London, SW7 2BP, UK.
[2]Department of Earth Science and Engineering, Imperial College London; London, SW7 2BP, UK.

[3]European Centre for Medium-Range Weather Forecast; Reading, RG2 9AX, UK.

[4]Department of Scientific Computing, Florida State University; Tallahassee, FL 32306-4120, USA.

*Corresponding author. Email: f.fang@imperial.ac.uk

**Abstract:** PM2.5 forecasting is crucial for public health, air quality management, and policy development. Traditional physics-based models are computationally demanding and slow to adapt to real-time conditions. Deep learning models show potential in efficiency but still suffer from accuracy loss over time due to error accumulation. To address these challenges, we propose a dual deep neural network (D-DNet) prediction and data assimilation system that efficiently integrates real-time observations, ensuring reliable operational forecasting. D-DNet excels in global operational forecasting for PM2.5 and AOD550, maintaining consistent accuracy throughout the entire year of 2019. It demonstrates notably higher efficiency than the Copernicus Atmosphere Monitoring Service (CAMS) 4D-Var operational forecasting system while maintaining comparable accuracy. This efficiency benefits ensemble forecasting, uncertainty analysis, and large-scale tasks.

**Main Text:** Accurately forecasting PM2.5 concentration (Particulate Matter with a diameter of 2.5 micrometers or smaller) is of paramount importance, given its significant impact on air quality and public health (*1, 2*). In general, atmospheric forecasting is a challenging task due to the complex and chaotic nature of the atmosphere system (*3–5*). In the context of PM2.5 forecasting, the complexity is compounded by the intricate interactions between various atmospheric processes, emissions, and depositions (*6–8*). Traditional forecasting methods, such as the atmospheric models in Copernicus Atmosphere Monitoring Service (CAMS), are based on the fundamental physical and chemical principles governing the emission, transformation, and transport of pollutants. The need to accurately represent numerous complex processes leads to significant computational demands (*9, 10*). Running these models in a timely manner, which is crucial for forecasting, requires access to high-performance computing resources (*11, 12*).

To address these challenges, the scientific community has explored using advanced neural networks as complementary to physics-based models (*13–15*). Neural network-based models can provide efficient forecasting with low computational expense and facilitate ensemble forecasting and uncertainty analysis for large-scale forecasting. Previous studies have successfully employed neural networks for medium-range weather forecasting, achieving notable success in both efficiency and accuracy (*14, 16, 17*). However, PM2.5 forecasting has received less attention. Nonetheless, one persistent challenge faced by current neural network-based atmospheric forecasting is the error accumulation arising from uncertainties (for example, initial conditions and emissions). This issue has been recognized in previous research but has yet to be comprehensively addressed (*14*). To bridge this gap and advance the field of PM2.5 forecasting, we introduce a novel approach, Dual Deep Neural Networks (D-DNet), for operational PM2.5 forecasting, which periodically integrates real-time observations to avoid error accumulations and enhance operational forecasting accuracy.

The Proposed D-DNet operational forecasting system consists of two distinct neural networks: PredNet for prediction and DANet for real-time updates during the data assimilation (DA) process. First, PredNet is developed to make initial forecasts based on historical input features. Leveraging the power of neural networks, the PredNet captures complex nonlinearities and patterns within the datasets. The well-trained PredNet provides computationally efficient forecasts that eliminate the need for resource-intensive physical simulations. Subsequently, DANet is trained to enhance the PredNet forecasts by assimilating the available observational data.

Compared to traditional operational forecasting methods that employ physics-based models and variational or ensemble-based DA systems, neural network-based approaches offer significant advantages (*16, 18, 19*). Neural networks are inherently data-driven, allowing them to capture hidden patterns and relationships that physics-based models may overlook. Our proposed D-DNet enables efficient and effective large-scale operational spatiotemporal forecasting. Furthermore, the real-time integration of observations into forecasts via the DANet ensures improved initial conditions, resulting in an overall improvement in operational forecasting performance.

To demonstrate the effectiveness of the proposed D-DNet operational forecasting system, we conducted a case study to forecast global PM2.5 concentrations and AOD550 (Aerosol Optical Depth at 550 nm) values. Our case study shows that the proposed D-DNet can achieve reliable operational forecasting of PM2.5 concentration and AOD550 values throughout the entire year of 2019. D-DNet also exhibited greater efficiency than the CAMS 4D-Var operational forecasting system.

**Operational forecasting with D-DNet**

In our fully deep neural network-based approach, we harness the strengths of neural networks for both the prediction and update stages of operational forecasting (Fig. 1). In the prediction phase, a neural network (PredNet) is trained using historical data to learn complex patterns and interconnections within physical processes. By utilizing multivariate input data, PredNet efficiently captures dependencies and interactions, resulting in a robust forecasting model. Its ability to uncover intricate patterns in large datasets enables PredNet to provide efficient and reliable forecasts for the target variable.

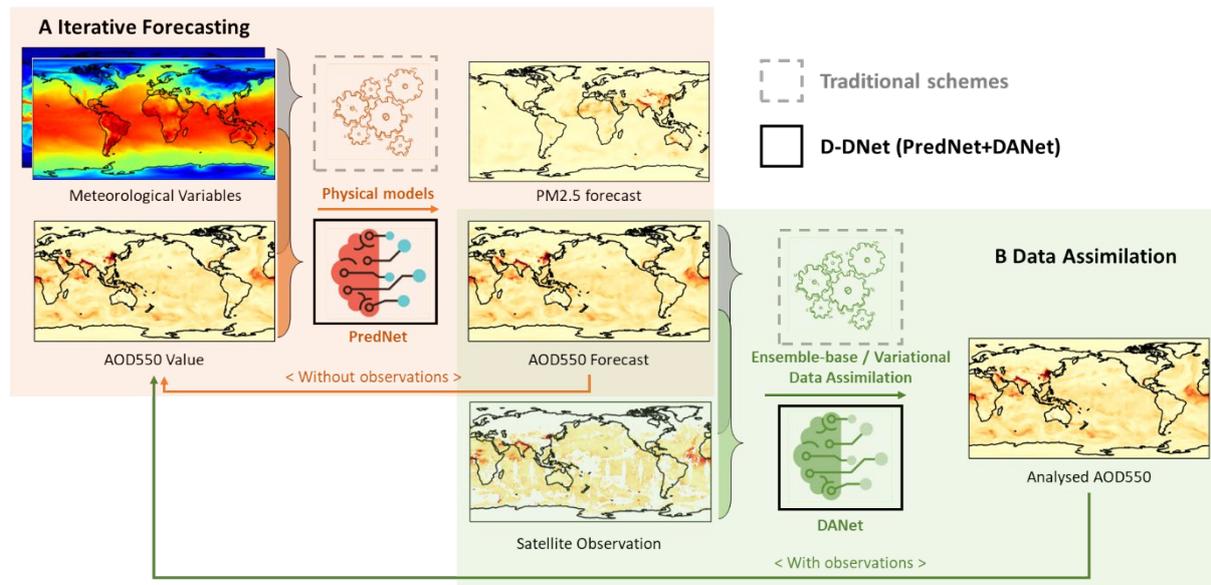

**Fig. 1. Workflow illustration of D-DNet.** The D-DNet framework is designed for operational forecasting and employs neural networks: one for prediction (PredNet) and one for data assimilation (DANet). (**A**) Iterative forecasting step: in this phase, meteorological variables and historical AOD550 values are input into PredNet to produce PM2.5 and AOD550 forecasts. The forecasting is conducted iteratively, with the forecasted AOD550 values from each step serving as inputs for the subsequent step. (**B**) Data assimilation step: DANet integrates observations into the PredNet forecasts to generate a refined result, referred to as the analysis result. This refined result is then used as input for the subsequent PredNet forecasting step. For comparative purposes, D-DNet components are marked with black squares, while corresponding components in traditional schemes are marked with dash grey squares.

In the DA step, neural networks play a crucial role in enhancing forecast accuracy through the timely integration of new observations. DANet is specially trained to identify discrepancies between PredNet forecasts and actual observations. This acquired knowledge from observations enables DANet to refine PredNet forecasts whenever new observations are available. The outcome of this refinement process, referred to as 'analysis results', represents the optimal estimation of the current state of the studied dynamic system, effectively integrating PredNet forecasts with the latest observations.

The D-DNet operational forecasting approach employs iterative forecasting and DA processes, where the analysis state variable from DA serves as the initial condition for subsequent forecasting cycles. This strategy mitigates model error accumulation, enhances the overall performance of operational forecasting, and ensures long-term stability (*20*). Traditional schemes, typically rely on physical models and utilize ensemble-based or variational DA techniques. These schemes often encounter challenges posed by nonlinearities, large-scale systems, and growing volumes of observations (*21*). In contrast, D-DNet demonstrates superiority in these areas and adapts more effectively to evolving dynamics. By leveraging the computational efficiency of neural networks, the D-DNet framework seamlessly integrates observations and generates real-time analysis results, thus overcoming persistent challenges faced by traditional forecasting methods.

## Results

We conducted global PM2.5 and AOD550 forecasting using the proposed D-DNet. The detrimental effects of PM2.5 pollution on human health and the environment have prompted the need for accurate and timely forecasting of PM2.5 concentrations (*1*, *2*). However, this forecasting encounters formidable challenges due to significant temporal and spatial variations, which are influenced by various factors, including weather conditions (i.e. winds, temperature, and precipitations), emission sources, and topographical features (*6–8*).

Our innovative operational forecasting model, D-DNet, offers $0.75° \times 0.75°$ forecasting with a temporal resolution of 3 hours, the same as the EAC4 (ECMWF Atmospheric Composition Reanalysis) dataset (details can be found in the supplementary material section 1), which is the latest global reanalysis dataset of atmospheric composition (*22*). EAC4 serves as the "ground truth" for model training and forecasting evaluation in this study. The emission data sources used as inputs in PredNet are derived from the CAMS-GLOB-ANT dataset, which offers monthly global emissions for 36 compounds (*23*). These emission data are then temporally distributed to each hour based on the CAMS TEMPOral profiles (CAMS-TEMPO) dataset, providing detailed temporal profiles for global emissions in atmospheric chemistry modeling (*24*). In addition to PM2.5 concentrations, we also forecasted AOD550 values, which are positively correlated with PM2.5 and can be measurable over large areas via satellite. The correlation between AOD550 and PM2.5 varies over time and space, influenced by meteorological factors like relative humidity and boundary layer height, as well as topography (*25*, *26*). Details of the correlation analysis among available variables can be found in the supplementary material section 2. For observations, we adopted AOD550 satellite observations obtained from the MOD08 and MYD08 datasets (*27*) (for details, see supplementary material section 1).

Detailed descriptions of PredNet and DANet in the D-DNet operational forecasting system are provided in the supplementary material section 3. Initially, PredNet was trained using the EAC4 reanalysis dataset, specifically PM2.5 concentrations, AOD550 values, meteorological conditions, and topography from 2011 to 2017. The trained PredNet was used to conduct 5-day-ahead forecasting in 2018. Subsequently, DANet was trained using the AOD550 forecasts from PredNet along with satellite observations from the MODIS dataset in 2018 to correct PredNet forecasts. With the trained PredNet and DANet, we conducted operational forecasting for the entire year of 2019. The forecasting began at 00:00 UTC (Coordinated Universal Time) on 1 January 2019, starting with an initial condition identical to that in the CAMS 4D-Var forecasts, and continued throughout the year. Implementation details of the operational forecasting are provided in the supplementary material section 3. Evaluation results for PredNet and DANet are presented in the supplementary material sections 5 and 6.

*PM2.5 and AOD550 forecast evaluation*

We conducted a comparative analysis between D-DNet and the physics-based CAMS 4D-Var system for the operational forecasting of PM2.5 and AOD550. In this comparison, both D-DNet and CAMS 4D-Var systems deliver operational forecasts with the same DA frequency of 12 hours. D-DNet and CAMS 4D-Var operational forecasts were evaluated against the EAC4 reanalysis dataset. For consistency, the CAMS 4D-Var operational forecasts, which have a 0.4° × 0.4° grid resolution, were down-sampled to match the EAC4 dataset with a 0.75° × 0.75° gird resolution. For PM2.5 operational forecasting in 2019, D-DNet forecasts exhibited a lower mean Root Mean Square Error (RMSE) and a slightly higher mean correlation coefficient (R) compared to CAMS 4D-Var forecasts. Similarly, for AOD550 operational forecasting over the same period, D-DNet forecasts demonstrated improved accuracy with a lower mean RMSE value and a higher mean R value compared to CAMS 4D-Var. Detailed values supporting these findings are presented in Table 1.

**Table 1. Mean RMSE and R for PM2.5 and AOD550 forecasts** using D-DNet and the CAMS 4D-Var operational forecasting system.

|                        | RMSE.mean | R.mean |
|------------------------|-----------|--------|
| PM2.5 / D-DNet         | 18.04     | 0.73   |
| PM2.5 / CAMS 4D-Var    | 19.93     | 0.70   |
| AOD550 / D-DNet        | 0.07      | 0.87   |
| AOD550 / CAMS 4D-Var   | 0.08      | 0.75   |

Figure 2 presents the RMSE and R metrics throughout the forecasting period, covering the entire year of 2019. D-DNet demonstrates lower RMSE values and higher R values compared to the CAMS 4D-Var system in ~78% of all the forecasting time steps for both PM2.5 and AOD550. Additionally, RMSE spikes observed in both models suggest that certain meteorological conditions or events occasionally pose challenges to forecasting accuracy.

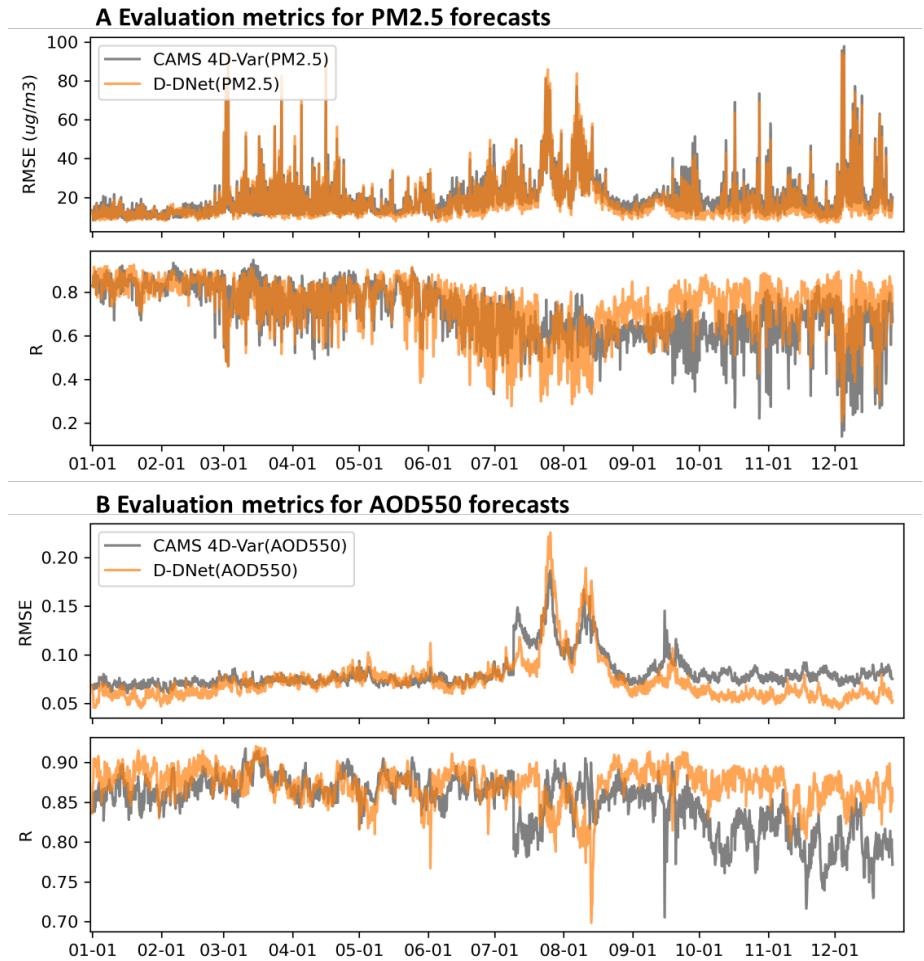

**Fig. 2. Global operational forecast evaluation. (A)** Evaluation metrics (RMSE and R) for PM2.5 forecasts compared to the "ground truth" in 2019. **(B)** Evaluation metrics for AOD550 forecasts compared to the "ground truth" in 2019. Operational forecasts from D-DNet are compared with those from the CAMS 4D-Var system, a renowned system for global atmospheric composition forecasting. Both D-DNet and the CAMS 4D-Var system deliver operational forecasts with the same DA frequency of 12 hours.

To further evaluate D-DNet and the CAMS 4D-Var system, we conducted a Cumulative Accuracy Profile (CAP) analysis based on the distribution of RMSE and R metrics across the forecasting period (Fig. 3). Specifically, we calculated the proportion of forecast instances where RMSE is lower than a specific threshold $RMSE_i$ $P(RMSE<RMSE_0)$ and where R is larger than a specific threshold $R_i$ $P(R>R_i)$. This set of threshold values $RMSE_i$ and $R_i$ covers all observed RMSE and R values for both models. Further details and calculations of $P(RMSE<RMSE_i)$ and $P(R>R_i)$ are provided in the supplementary material section 4.

Figure 3, A and B, display the cumulative distribution function (CDF) for PM2.5 in terms of RMSE and R. D-DNet shows a consistently higher CDF for both RMSE and R across the majority of the threshold values, suggesting better accuracy compared to the CAMS 4D-Var system. Figure 3, C and D, display the CDF for AOD550. Similar to the PM2.5 results, Figure 3C indicates that D-DNet maintains a higher frequency of lower RMSE values for AOD550 forecasting, while Figure 3D illustrates that D-DNet more often attains higher R values than the CAMS 4D-Var system. The convergence of both models at the lower and upper tails of the

RMSE and R thresholds suggests that the performance differential is most pronounced at intermediate thresholds. This suggests that while both models may perform similarly under extremely favorable or unfavorable conditions, D-DNet offers improved accuracy and reliability under typical operational conditions.

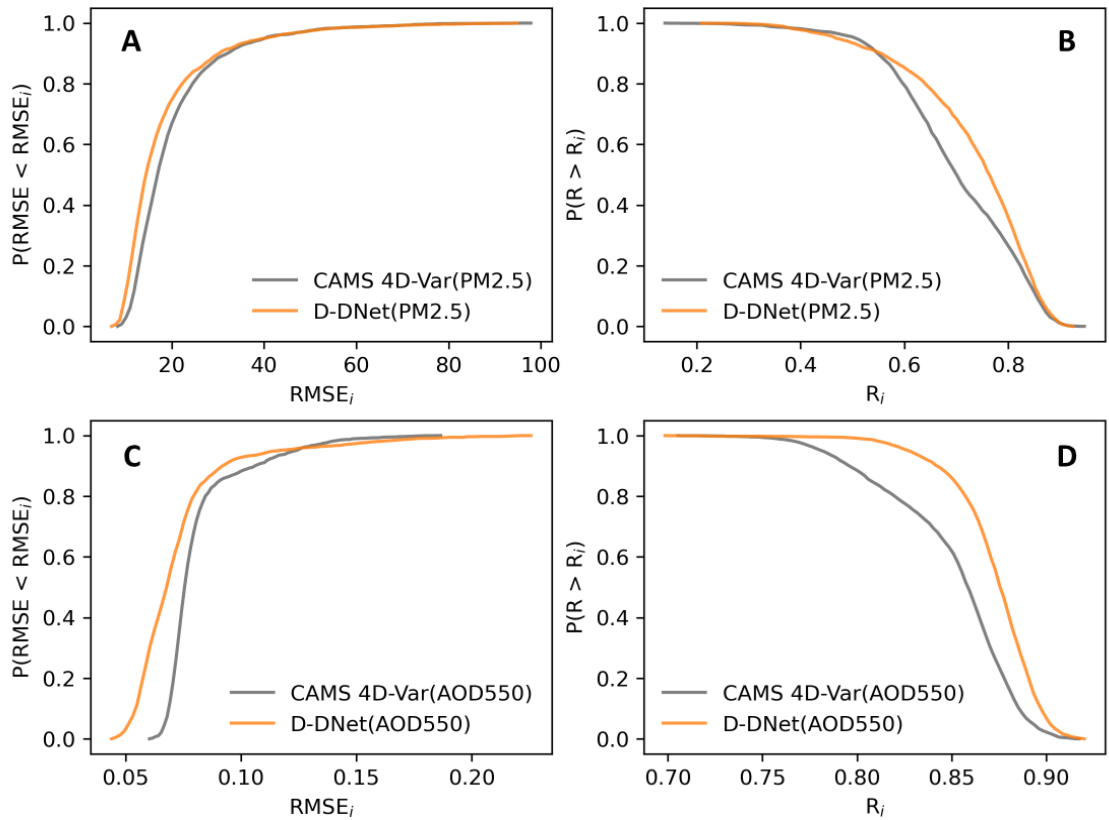

**Fig. 3. Cumulative Accuracy Profile (CAP) analysis** of D-DNet and the CAMS 4D-Var system for operational PM2.5 and AOD550 forecasts.

For a more comprehensive comparison and in-depth analysis, we present the spatial distribution of PM2.5 concentrations from D-DNet forecasts, CAMS 4D-Var forecasts, and EAC4 reanalysis at 18:00 UTC on different seasons: 1 February, 1 June, and 1 October 2019 (Fig. 4). The bottom two rows represent the forecast errors for D-DNet and the CAMS 4D-Var system, with grey indicating minimal errors, blue indicating underestimation, and red indicating overestimation of the PM2.5 concentration compared to EAC4. Our findings reveal that D-DNet successfully captures both the spatial patterns and seasonal fluctuations in global PM2.5 concentrations. For further comparison, we included regional evaluations and analyses for these forecasts in the supplementary material section 7. It is shown that D-DNet also successfully forecasts relatively high PM2.5 concentrations in Sub-Saharan, Northern India, and Northern China Plain. Across most areas, D-DNet provides relatively accurate PM2.5 forecasts with low errors. Comparatively, the CAMS 4D-Var system generally underestimates PM2.5 concentrations on a global scale in October.

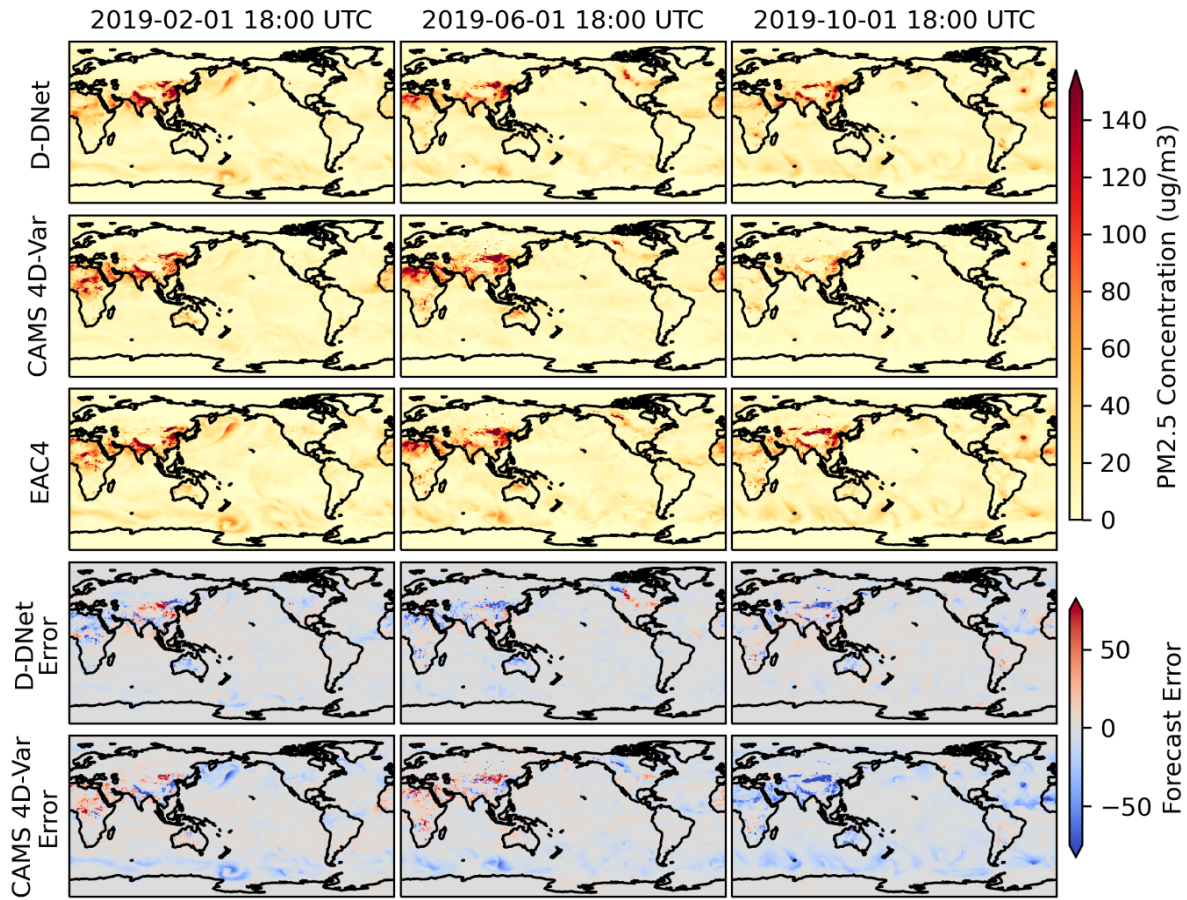

**Fig. 4. Spatial distribution of PM2.5 concentrations** from the D-DNet forecasts, CAMS 4D-Var forecasts, and EAC4 reanalysis ("ground truth") on 1 February, 1 June, and 1 October 2019, as well as corresponding forecasting errors.

Forecasts for AOD550 at the same time points are displayed in Fig. 5. D-DNet forecasts closely match "ground truth" with minimal errors across the majority area. Comparatively, the CAMS 4D-Var system shows a tendency to overestimate (reddish) and occasionally underestimate (blue) AOD550 values. Overall, D-DNet consistently provides reliable AOD550 forecasts across all examined dates. In regional assessments of AOD550 (refer to the supplementary material section 7), it reveals that D-DNet delivers better forecasts for the majority of regions compared to the CAMS 4D-Var system.

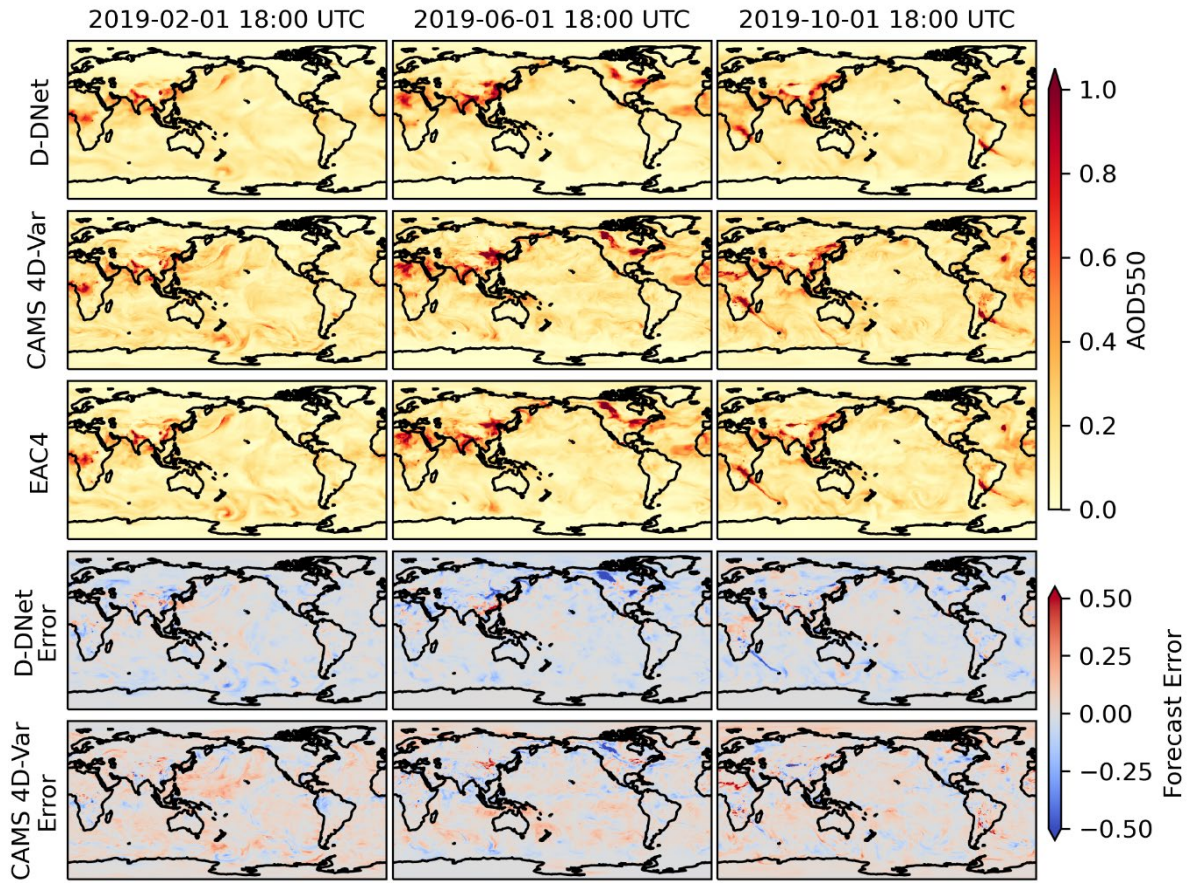

**Fig. 5. Spatial distribution of AOD550** from the D-DNet forecasts, CAMS 4D-Var forecasts, and EAC4 reanalysis ("ground truth") on 1 February, 1 June, and 1 October 2019, as well as corresponding forecasting errors.

We also evaluated the forecasting accuracy of D-DNet against a neural network baseline, specifically PredNet without DA, which serves as the benchmark for existing neural network-based forecasting models. This comparison underscores the importance of incorporating DANet into the proposed D-DNet, a scheme that, to our knowledge, has not been explored in previous studies yet. A detailed comparison and analysis are presented in the supplementary material section 8. As the forecast time extends, PredNet forecasts tend to accumulate errors and experience accuracy loss. In contrast, the D-DNet model, which periodically incorporates satellite observations, demonstrates consistent performance throughout the entire forecasting period. This improvement in D-DNet performance can be attributed to its advanced integration of real-time satellite data, which likely maintains its accuracy and prevents error divergence over time. These findings highlight the effectiveness of D-DNet in operational forecasting for PM2.5 and AOD550.

*Computational efficiency*

D-DNet is much faster than the CAMS 4D-Var system for operational PM2.5 forecasting, despite differences in devices and spatial and temporal resolutions. In this study, D-DNet runs on a single NVIDIA RTX A4000 GPU (16G) device. It takes approximately 40 seconds to generate a 5-day forecast with a spatial resolution of $0.75° \times 0.75°$ and a temporal resolution of 3 hours. For the whole workflow including initial analysis and 5-day forecast, D-DNet only takes 41 seconds. For comparison, CAMS runs on ECMWF's ATOS HPC which is in Bologna,

Italy. It consists of 7680 nodes (128 cores and 256GB of memory per node) and the CPU architecture is AMD Rome (2.5 GHZ). CAMS forecast runs on 32 nodes (using 32*128=4096 cores). The forecast resolution is TL511 (about 40 km), with a time step of 900 seconds and 137 model levels. Since the update to the last Integrated Forecasting System (IFS) cycle 48r1 in June 2023, the 5-day operational global forecast itself has an average run time of 832 seconds (with a standard deviation of 40 seconds). Considering the whole workflow steps involved in the operational production of each 5-day forecast, which includes fetching observations, pre-screening observations, 4D-Var analysis, forecasting, and product dissemination, the total time would be around 2 hours. In summary, the D-DNet provides much more efficient operational forecasting while maintaining a level of accuracy comparable to the CAMS 4D-Var system.

**Conclusion**

In this study, we introduced the dual deep neural network (D-DNet) as a cutting-edge solution for operational forecasting. D-DNet seamlessly integrates real-time observations and model forecasts, demonstrating its ability to generate robust and reliable operational global PM2.5 forecasts throughout the entire year of 2019. Our results show that the D-DNet delivers PM2.5 and AOD550 forecasts more closely aligned with the EAC4 reanalysis dataset than those from the CAMS 4D-Var system. Its exceptional computational efficiency facilitates real-time DA at a high spatial resolution, making it well-suited for addressing large-scale forecasting challenges. D-DNet effectively addresses the computational demands typical of traditional physics-based models and mitigates the error accumulation found in current neural network-based approaches. The success of D-DNet in PM2.5 forecasting opens exciting prospects for advancements in the operational forecasting domain. Embracing the synergy between deep neural networks, real-time observations, and physical-based models, the proposed D-DNet promises accurate, timely, and efficient forecasts. This empowers decision-makers across various domains and marks a significant step forward in the pursuit of enhanced forecasting accuracy and efficacy.

**References and Notes**

1.  Y. Xie, H. Dai, H. Dong, T. Hanaoka, T. Masui, Economic Impacts from PM 2.5 Pollution-Related Health Effects in China: A Provincial-Level Analysis. *Environ Sci Technol* **50**, 4836–4843 (2016).

2.  X. Bu, Z. Xie, J. Liu, L. Wei, X. Wang, M. Chen, H. Ren, Global PM2.5-attributable health burden from 1990 to 2017: Estimates from the Global Burden of disease study 2017. *Environ Res* **197**, 111123 (2021).

3.  P. Bauer, P. D. Dueben, T. Hoefler, T. Quintino, T. C. Schulthess, N. P. Wedi, The digital revolution of Earth-system science. *Nat Comput Sci* **1**, 104–113 (2021).

4.  C. J. Walcek, N. M. Aleksic, A simple but accurate mass conservative, peak-preserving, mixing ratio bounded advection algorithm with FORTRAN code. *Atmos Environ* **32**, 3863–3880 (1998).

5.  Z. Wang, W. Sha, H. Ueda, Numerical modeling of pollutant transport and chemistry during a high-ozone event in northern Taiwan. *Tellus B: Chemical and Physical Meteorology* **52**, 1189–1205 (2000).

6.  Q. Zhang, Y. Zheng, D. Tong, M. Shao, S. Wang, Y. Zhang, X. Xu, J. Wang, H. He, W. Liu, Y. Ding, Y. Lei, J. Li, Z. Wang, X. Zhang, Y. Wang, J. Cheng, Y. Liu, Q. Shi, L. Yan, G. Geng, C. Hong, M. Li, F. Liu, B. Zheng, J. Cao, A. Ding, J. Gao, Q. Fu, J. Huo, B. Liu, Z. Liu, F. Yang, K. He, J. Hao, Drivers of improved PM 2.5 air quality in China from 2013 to 2017. *Proceedings of the National Academy of Sciences* **116**, 24463–24469 (2019).

7.  F. Liang, Q. Xiao, K. Huang, X. Yang, F. Liu, J. Li, X. Lu, Y. Liu, D. Gu, The 17-y spatiotemporal trend of PM $_{2.5}$ and its mortality burden in China. *Proceedings of the National Academy of Sciences* **117**, 25601–25608 (2020).

8.  Z. Chen, D. Chen, C. Zhao, M. Kwan, J. Cai, Y. Zhuang, B. Zhao, X. Wang, B. Chen, J. Yang, R. Li, B. He, B. Gao, K. Wang, B. Xu, Influence of meteorological conditions on PM2.5 concentrations across China: A review of methodology and mechanism. *Environ Int* **139**, 105558 (2020).

9.  S. Ravuri, K. Lenc, M. Willson, D. Kangin, R. Lam, P. Mirowski, M. Fitzsimons, M. Athanassiadou, S. Kashem, S. Madge, R. Prudden, A. Mandhane, A. Clark, A. Brock, K. Simonyan, R. Hadsell, N. Robinson, E. Clancy, A. Arribas, S. Mohamed, Skilful precipitation nowcasting using deep generative models of radar. *Nature* **597**, 672–677 (2021).

10. M. G. Jacox, M. A. Alexander, D. Amaya, E. Becker, S. J. Bograd, S. Brodie, E. L. Hazen, M. Pozo Buil, D. Tommasi, Global seasonal forecasts of marine heatwaves. *Nature* **604**, 486–490 (2022).

11. D. Kochkov, J. A. Smith, A. Alieva, Q. Wang, M. P. Brenner, S. Hoyer, Machine learning–accelerated computational fluid dynamics. *Proceedings of the National Academy of Sciences* **118** (2021).

12. M. Cheng, F. Fang, I. M. Navon, J. Zheng, X. Tang, J. Zhu, C. Pain, Spatio-Temporal Hourly and Daily Ozone Forecasting in China Using a Hybrid Machine Learning Model: Autoencoder and Generative Adversarial Networks. *J Adv Model Earth Syst* **14**, 1–26 (2022).

13. J. Pathak, S. Subramanian, P. Harrington, S. Raja, A. Chattopadhyay, M. Mardani, T. Kurth, D. Hall, Z. Li, K. Azizzadenesheli, P. Hassanzadeh, K. Kashinath, A. Anandkumar, FourCastNet: A Global Data-driven High-resolution Weather Model using Adaptive Fourier Neural Operators. arXiv:2202.11214v1 [physics.ao-ph] (2022).

14. K. Bi, L. Xie, H. Zhang, X. Chen, X. Gu, Q. Tian, Accurate medium-range global weather forecasting with 3D neural networks. *Nature* **619**, 533–538 (2023).

15. R. Lam, A. Sanchez-Gonzalez, M. Willson, P. Wirnsberger, M. Fortunato, F. Alet, S. Ravuri, T. Ewalds, Z. Eaton-Rosen, W. Hu, A. Merose, S. Hoyer, G. Holland, O. Vinyals, J. Stott, A. Pritzel, S. Mohamed, P. Battaglia, Learning skillful medium-range global weather forecasting. *Science* **382**, 1416–1421 (2023).

16. M. Reichstein, G. Camps-Valls, B. Stevens, M. Jung, J. Denzler, N. Carvalhais, Prabhat, Deep learning and process understanding for data-driven Earth system science. *Nature* **566**, 195–204 (2019).

17. H. Wu, H. Zhou, M. Long, J. Wang, Interpretable weather forecasting for worldwide stations with a unified deep model. *Nat Mach Intell* **5**, 602–611 (2023).

18. Y. Qi, Q. Li, H. Karimian, D. Liu, A hybrid model for spatiotemporal forecasting of PM 2.5 based on graph convolutional neural network and long short-term memory. *Science of the Total Environment* **664**, 1–10 (2019).

19. M. Leutbecher, Ensemble size: How suboptimal is less than infinity? *Quarterly Journal of the Royal Meteorological Society* **145**, 107–128 (2019).

20. A. Gettelman, A. J. Geer, R. M. Forbes, G. R. Carmichael, G. Feingold, D. J. Posselt, G. L. Stephens, S. C. van den Heever, A. C. Varble, P. Zuidema, The future of Earth system prediction: Advances in model-data fusion. *Sci Adv* **8**, 3488 (2022).

21. T. C. Vance, T. Huang, K. A. Butler, Big data in Earth science: Emerging practice and promise. *Science* **383** (2024).

22. A. Inness, M. Ades, A. Agustí-Panareda, J. Barré, A. Benedictow, A.-M. Blechschmidt, J. J. Dominguez, R. Engelen, H. Eskes, J. Flemming, V. Huijnen, L. Jones, Z. Kipling, S. Massart, M. Parrington, V.-H. Peuch, M. Razinger, S. Remy, M. Schulz, M. Suttie, The CAMS reanalysis of atmospheric composition. *Atmos Chem Phys* **19**, 3515–3556 (2019).

23. A. Soulie, C. Granier, S. Darras, N. Zilbermann, T. Doumbia, M. Guevara, J.-P. Jalkanen, S. Keita, C. Liousse, M. Crippa, D. Guizzardi, R. Hoesly, S. J. Smith, Global anthropogenic emissions (CAMS-GLOB-ANT) for the Copernicus Atmosphere Monitoring Service simulations of air quality forecasts and reanalyses. *Earth Syst Sci Data* **16**, 2261–2279 (2024).

24. C. Granier, S. Darras, H. D. van der Gon, J. Doubalova, N. Elguindi, B. Galle, M. Gauss, M. Guevara, J.-P. Jalkanen, J. Kuenen, C. Liousse, B. Quack, D. Simpson, K. Sindelarova, "The Copernicus Atmosphere Monitoring Service global and regional emissions" (2019); https://doi.org/10.24380/d0bn-kx16.

25. Q. He, M. Wang, S. H. L. Yim, The spatiotemporal relationship between PM2.5 and aerosol optical depth in China: influencing factors and implications for satellite PM2.5 estimations using MAIAC aerosol optical depth. *Atmos Chem Phys* **21**, 18375–18391 (2021).

26. J. Handschuh, T. Erbertseder, M. Schaap, F. Baier, Estimating PM2.5 surface concentrations from AOD: A combination of SLSTR and MODIS. *Remote Sens Appl* **26**, 100716 (2022).

27. R. C. Levy, L. A. Remer, D. Tanré, S. Mattoo, Y. J. Kaufman, "ALGORITHM FOR REMOTE SENSING OF TROPOSPHERIC AEROSOL OVER DARK TARGETS FROM MODIS: Collections 005 and 051: Revision 2; Feb 2009" (2009).

**Acknowledgments:** The authors acknowledge Dr. Peter Dueben for his in-depth perspicacious comments that contributed to improving the presentation of this paper.

**Funding:** The authors are grateful to the sponsors of the Resource Geophysics Academy, Imperial College London for supporting this research. Additional funding is from the Engineering and Physical Sciences Research Council (EPSRC) (EP/X029093/1, AI-Respire (EP/Y018680/1), ECO-AI (EP/Y005732/1), MAGIC (EP/N010221/1), and INHALE (EP/T003189/1) in the UK.

**Author contributions:**

Conceptualization: F.F., and S.C.

Data curation: S.C, F.F, V.-H.P., and M.A.

Methodology: S.C., and F.F.

Investigation: S.C, F.F, V.-H.P., M.A., and M.I.N.

Visualization: S.C.

Funding acquisition: Y.W., and F.F.

Project administration: F.F., and Y.W.

Supervision: F.F., and Y.W.

Validation: S.C, F.F, V.-H.P., M.A., M.I.N., and Y.W.

Writing – original draft: S.C.

Writing – review & editing: S.C., F.F., V.-H.P., M.A., M.I.N., and Y.W.

**Competing interests:** Authors declare that they have no competing interests.

**Data and materials availability:** D-DNet's code and trained weights are publicly accessible on GitHub at https://github.com/SJ-CAI/D-DNet. The EAC4 datasets used for training, evaluating, and testing D-DNet, specifically for forecasting PM2.5 concentrations and AOD550 values, are from the Copernicus Atmosphere Monitoring Service (CAMS) Atmosphere Data Store (ADS) [https://www.ecmwf.int/en/forecasts/dataset/cams-global-reanalysis]. Additionally, emission sources and related temporal profiles were obtained from [https://ads.atmosphere.copernicus.eu/cdsapp#!/dataset/cams-global-emission-inventories]. Observational data utilized in Data Assimilation (DA) were sourced from MODIS products MOD08 and MYD08, available at NASA Earth Science Data [https://www.earthdata.nasa.gov/]. To benchmark our D-DNet model, we utilized operational forecasts from the CAMS 4D-Var system for PM2.5 concentrations and AOD550 values, accessible at [https://www.ecmwf.int/en/forecasts/dataset/cams-global-atmospheric-composition-forecasts]. All datasets employed in this study are publicly available and can be used for research purposes.

**Supplementary Materials**

Materials and Methods

Supplementary Text

Figs. S1 to S18

Tables S1 to S2

References 28–40

Supplementary Materials for

**Advancing operational PM2.5 forecasting with dual deep neural networks (D-DNet)**

Shengjuan Cai, Fangxin Fang, Vincent-Henri Peuch, Mihai Alexe, Ionel Michael Navon, Yanghua Wang

Corresponding author: Fangxin Fang (f.fang@imperial.ac.uk)

**The PDF file includes:**

# Contents

**Materials and Methods**

*1. Datasets*

*EAC4 dataset* (22): The EAC4 (ECMWF Atmospheric Composition Reanalysis) dataset is employed as the reference or "ground truth" of atmospheric composition states in our study. This dataset is used both for model training and validation for our forecasting results. EAC4 was produced by optimally integrating physical simulations and re-processed observations from many satellites for the last few decades. This integration results in a consistent and high-quality dataset that can be used for scientific studies and trend analysis. The integration is executed using the four-dimensional variational data assimilation method (4DVar) in CY42R1 of ECMWF's Integrated Forecast System (IFS) with an assimilation window of 12 hours. EAC4 has a resolution of approximately $0.75° \times 0.75°$ with a sub-daily and monthly frequency.

*CAMS-GLOB-ANT dataset* (23): The CAMS-GLOB-ANT dataset, developed as part of the Copernicus Atmosphere Monitoring Service (CAMS), offers monthly global emissions data for 36 compounds. These compounds include key air pollutants and greenhouse gases across 17 sectors. Covering the 2000-2023 period at a fine spatial resolution of $0.1° \times 0.1°$. This dataset aids in forecasting atmospheric composition and provides essential insights into the impact of anthropogenic emissions on the environment. We incorporate emissions of Black Carbon (BC) and Organic Carbon (OC) from this dataset into both the training and prediction processes of our forecasting model, ensuring consideration of their impact on PM2.5 and Aerosol Optical Depth at 550 nanometers (AOD550) forecasting.

*CAMS-TEMPO dataset (emission temporal profile maps)* (28): The Copernicus Atmosphere Monitoring Service TEMPOral profiles (CAMS-TEMPO) dataset provides detailed temporal profiles for global and European emissions in atmospheric chemistry modeling. It includes essential data on air pollutants and greenhouse gases, providing grided weight factors on a monthly, daily, weekly, and hourly basis for different anthropogenic sources. We use this temporal profile to distribute monthly emissions from the CAMS-GLOB-ANT dataset to an hourly scale, enhancing the accuracy of hourly PM2.5 and AOD550 forecasting.

*MOD08 and MYD08 datasets (satellite observations)* (27): MOD08 and MYD08, both part of NASA's MODIS (Moderate Resolution Imaging Spectroradiometer) data collection, provide daily information about atmospheric properties. These datasets include data on atmospheric aerosols, water vapor, and cloud properties, derived from observations made by the Terra and Aqua satellites, respectively. In this study, we use AOD550 satellite observations from MOD08 and MYD08 datasets to enhance model forecasts through data assimilation (DA) techniques. The analysis AOD550 values, obtained by integrating numerical simulations and satellite AOD550 observations, are used as initial conditions to constrain 5-day forecasting.

*CAMS global atmospheric composition forecasts dataset (baseline)* (29): CAMS provides global forecasts for more than 50 chemical species and 7 different types of aerosols, using numerical atmospheric models and DA. CAMS offers global 5-day forecasts twice daily at 00 and 12 UTC. CAMS stands as a renowned provider of global atmospheric composition forecasts, supporting air quality monitoring and environmental research. In our study, this dataset serves as a baseline of physics-based atmospheric composition forecasting, facilitating comprehensive comparisons with our proposed neural network-based forecasting results. To ensure a fair comparison, we adopted the same configuration and initial conditions as CAMS for our forecasting model. We initiated our 5-day forecasts daily at 00 and 12 UTC, utilizing identical initial PM2.5 concentrations and AOD values as those found in the CAMS dataset. CAMS dataset has a spatial resolution of $0.4° \times 0.4°$ and a temporal resolution of 1-hour in a single level. To align with the EAC4 dataset for comparison or analysis, we down-sampled the CAMS dataset to match the same spatial ($0.75° \times 0.75°$) and temporal (3-hour) resolution as the EAC4 dataset.

Note that we utilized the CAMS global atmospheric composition forecasts dataset to conduct two benchmarks for our study.

*Benchmark 1: PredNet vs. CAMS 5-day forecasts*

The first benchmark involves a comparison between the 5-day forecasts generated by our PredNet and the numerical atmospheric composition models used in CAMS. In this scenario, no DA is employed in either set of results. We refer to these forecasts as "PredNet 5-day forecasts" and "CAMS 5-day forecasts" respectively. The purpose of this comparison is to demonstrate the improvements in forecasting accuracy achieved through the use of a neural network-based approach.

*Benchmark 2: D-DNet vs. CAMS operational forecasts*

The second benchmark compares the operational forecasts produced by our proposed D-DNet with those by the CAMS 4D-Var system. In this comparison, forecasting and DA processes are involved in both models. D-DNet has PredNet for forecasting and DANet for DA, while the CAMS 4D-Var system utilizes its numerical atmospheric models for forecasting and 4D-Var for DA. To ensure a fair comparison, both models were initialized with the same initial conditions and employed the same DA frequency. We refer to these forecasts as "D-DNet operational forecasts" and "CAMS 4D-Var operational forecasts" respectively. This comparison aims to demonstrate the improvements in operational forecasting accuracy achieved by incorporating neural networks in both forecasting and DA processes.

## *2. Correlation analysis*

We conducted a correlation analysis among available variables (Fig. S1). The heatmap analysis indicates that AOD550, BCAOD550, and OMAOD550 have the most significant positive correlations with PM2.5, with AOD550 showing the highest observed correlation of 0.60. This indicates that AOD550 is strongly associated with PM2.5 concentrations, consistent with the established understanding that both variables are indicators of particulate matter in the atmosphere *(30)*. The other variables show weak to negligible linear relationships with PM2.5, indicating that their relationships may be non-linear or influenced by other factors.

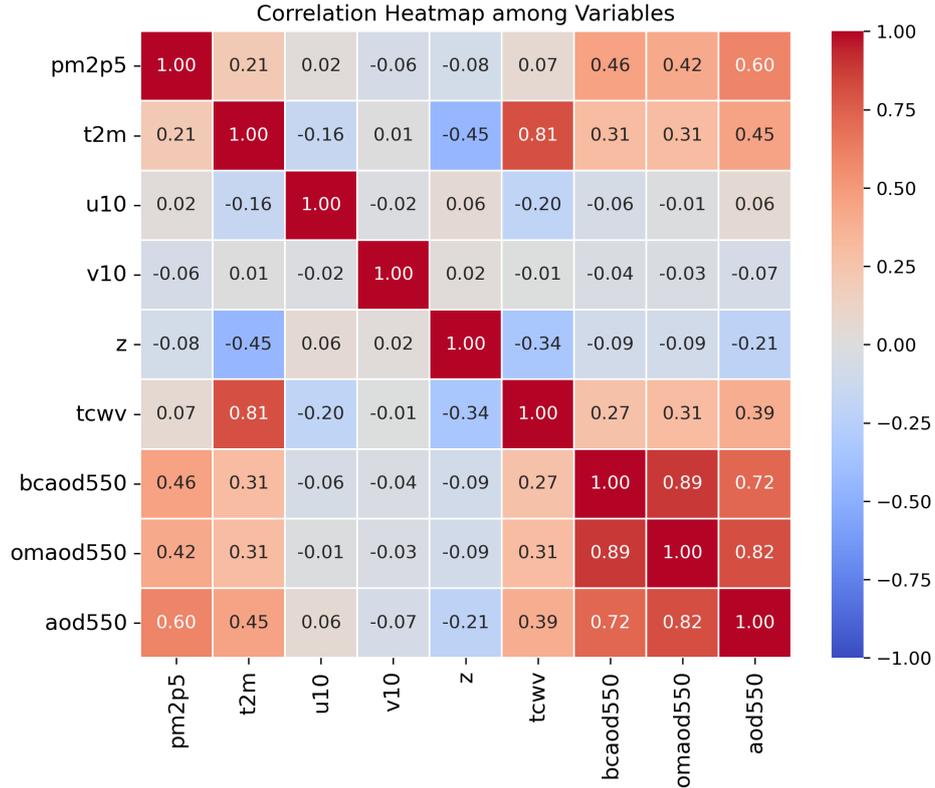

**Fig. S1. Correlation analysis among available variables** ('pm2p5': PM2.5 concentration; 't2m': Temperature at 2 meters; 'u10': Eastward Wind at 10 meters; 'v10': Northward Wind at 10 meters; 'z': Geopotential; 'tcwv': Total Column Water Vapour; 'bcaod550': Black Carbon Aerosol Optical Depth at 550 nm; 'omaod550': Organic Matter Aerosol Optical Depth at 550 nm; 'aod550': Aerosol Optical Depth at 550 nm)

*3. D-DNet model*

The D-DNet consists of two deep neural networks for iterative spatiotemporal forecasting (PredNet) and data assimilation (DANet). Initially, PredNet is trained for forecasting during the period [T0, T1] and subsequently utilized to generate the training dataset for DANet during the period [T1, T2]. Once both PredNet and DANet are trained, they are employed for operational forecasting in subsequent periods when t > T2. This workflow is illustrated in Fig. S2.

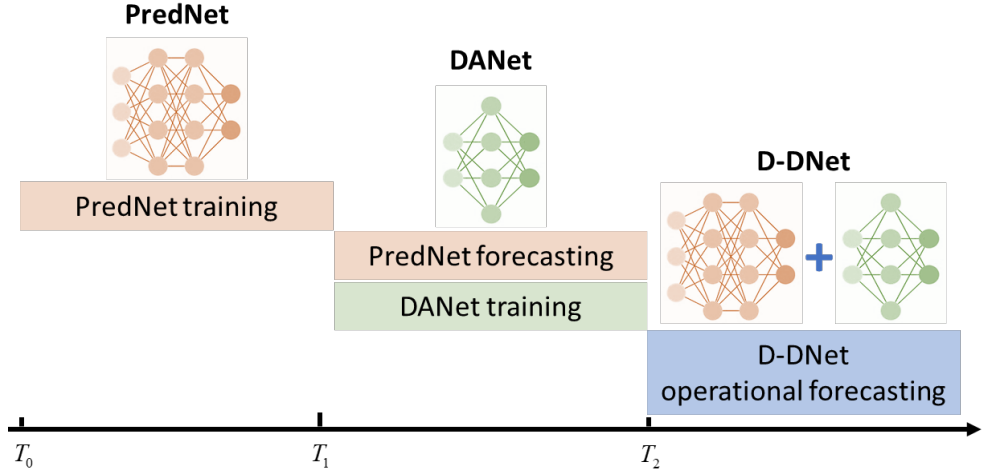

**Fig. S2. Implementation setups of D-DNet for operational forecasting**, in which PredNet and DANet are used for efficient spatiotemporal forecasting and data assimilation, respectively.

## *2.1 PredNet*

A PredNet is designed to forecast state variables from historical state variables and other influencing factors. In this study, the Convolutional Long Short-Term Memory (ConvLSTM) neural network is chosen as the primary architecture in the PredNet. The ConvLSTM network excels in spatiotemporal forecasting tasks because of its ability to capture both spatial and temporal dependencies effectively *(31, 32)*. The ConvLSTM network combines convolutional operators with LSTM architecture *(33, 34)*. The convolutional operators are used to extract spatial patterns and features from the input data, considering the interactions between neighboring locations. The LSTM architecture is employed to capture the temporal dynamics and long-term dependencies present in the studied system. By combining these two components, the ConvLSTM network enables the model to effectively learn and represent the complex relationships between future state variables, historical state variables, and other influencing factors. In this study, 4 ConvLSTM layers with different kernel sizes, $7 \times 7$, $5 \times 5$, $3 \times 3$, and $1 \times 1$, are used to capture spatial correlations and physical dynamics at different scales. The detailed model structure for PredNet is presented in Table S1.

This PredNet is trained on the EAC4 atmospheric composition reanalysis dataset, serving as a surrogate model for PM2.5 and AOD550 forecasting. The inputs for this model include the current AOD550 value, as well as meteorological variables (temperature, U- and V-components of wind, humidity, geopotential), and emissions (BC and OC) at the next time step. Here, the future meteorological variables are provided by ECMWF's high resolution weather forecasts, and future emissions are obtained from the CAMS-GLOB-ANT and the CAMS-TEMPO datasets. The outputs of the model are predicted PM2.5 concentration and AOD 550 value for the next time step.

**Table S1. PredNet model structure**

| Model: PredNet | | |
| --- | --- | --- |
| Layer | Output Shape | Parameter Number |
| InputLayer | [(None, None, 241, 480, 8)] | 0 |
| ConvLSTM2D | [(None, None, 241, 480, 64)] | 903,424 |
| BatchNormalization | [(None, None, 241, 480, 64)] | 256 |
| ConvLSTM2D | [(None, None, 241, 480, 64)] | 819,456 |
| BatchNormalization | [(None, None, 241, 480, 64)] | 256 |
| ConvLSTM2D | [(None, None, 241, 480, 64)] | 295,168 |
| BatchNormalization | [(None, None, 241, 480, 64)] | 256 |
| ConvLSTM2D | [(None, None, 241, 480, 64)] | 33,024 |
| Conv3D | [(None, None, 241, 480, 2)] | 3,458 |
| Total parameters: 2,055,298 | | |
| Trainable parameters: 2,054,914 | | |
| Non-trainable parameters: 384 | | |

The training process for PredNet is a supervised regression problem, which can be mathematically represented as an optimization problem. In this process, the model parameter of PredNet is optimized to minimize the discrepancy between the model outputs and the corresponding labels:

$$\min_{\boldsymbol{\theta}_f} L\left(f(\boldsymbol{\varphi}_t, \mathbf{u}_{t+\delta t}, \boldsymbol{\theta}_f), \ \boldsymbol{\varphi}_{t+\delta t}\right), \quad t \in [T_0, T_1), \tag{1}$$

where L denotes the loss function. In this study, we used the Mean Square Error (MSE) as the loss function in the model training process. The Adam optimization algorithm is employed to update the parameters of the PredNet iteratively. $f$ represents the PredNet, a nonlinear forecast model; $\boldsymbol{\theta}_f$ represents its trainable parameters. $\boldsymbol{\varphi}_t$ and $\mathbf{u}_{t+\delta t}$ are inputs of PredNet; $\boldsymbol{\varphi}_t$ represents the historical state variable at time $t$; $\mathbf{u}_{t+\delta t}$ the auxiliary variable at $t + \delta t$, which is temporarily stationary or predictable. $\boldsymbol{\varphi}_{t+\delta t}$ represents the "ground truth" state variable at time $t + \delta t$, serving as the label in the training process. The time step of PredNet is represented by $\delta t$. All samples from $T_0$ to $T_1$ constitute the training dataset (Fig. S2).

Once the PredNet is trained, the forecast process can be directly conducted using the following equation:

$$\tilde{\boldsymbol{\varphi}}_{t+\delta t}^f = f(\tilde{\boldsymbol{\varphi}}_t^f, \mathbf{u}_{t+\delta t}, \boldsymbol{\theta}_f), \quad t > T_1, \tag{2}$$

where $\tilde{\boldsymbol{\varphi}}_t^f$ and $\tilde{\boldsymbol{\varphi}}_{t+\delta t}^f$ denotes the forecasted state variable at time $t$ and $t + \delta t$, respectively. This forecasting is conducted iteratively, where the state variable forecasted at time $t$ is used as input for PredNet in the subsequent forecasting step. The PredNet forecasts from $T_1$ to $T_2$ will be used as the training dataset for DANet (Fig. S2).

Like other physical models, PredNet exhibits model errors due to its inability to perfectly represent all possible physical processes:

$$\boldsymbol{\varphi}_{t+\delta t} = \tilde{\boldsymbol{\varphi}}_{t+\delta t}^{f} + \boldsymbol{\varepsilon}_{t+\delta t}^{f}, \tag{3}$$

where $\boldsymbol{\varepsilon}_{t+\delta t}^{f}$ represents the forecast error at time $t+\delta t$. This forecast error accumulates in the iterative forecasting processes. To tickle this issue, we proposed to refine model forecasts and reduce forecast errors by assimilating observations using a DANet.

## 2.2 DANet

A DANet is designed to refine model forecasts and reduce forecast errors through DA when observations are available. DA is a technique used to obtain an optimal estimate of the current state of the studied dynamic system by combining observations with model forecasts *(35–37)*. Traditional DA methods, such as Kalman filter, variational methods, and ensemble-based methods, are computationally demanding for large-scale problems and struggle to capture the complexities of nonlinear and high-dimensional systems accurately *(38–40)*. To overcome these limitations, a neural network, DANet is proposed to conduct the DA process. In this study, 3 ConvLSTM layers with kernel sizes of $5 \times 5$, $3 \times 3$, and $1 \times 1$ are adopted as the main component of DANet, maintaining consistency with the PredNet throughout the entire operational forecasting process. The detailed model structure for DANet is presented in Table S2.

Our DANet assimilates daily AOD550 satellite observations from MOD08 and MYD08 datasets to AOD550 forecasts. The inputs to the DANet include AOD550 forecasts and the discrepancies between AOD550 forecasts and AOD550 satellite observations. The DANet outputs the discrepancies between AOD550 forecasts and the "ground truth". These DANet outputs are further used to update AOD550 forecasts.

**Table S2. DANet model structure**

| Layer | Output Shape | Parameter Number |
|---|---|---|
| InputLayer | [(None, None, 241, 480, 2)] | 0 |
| ConvLSTM2D | [(None, None, 241, 480, 64)] | 422,656 |
| BatchNormalization | [(None, None, 241, 480, 64)] | 256 |
| ConvLSTM2D | [(None, None, 241, 480, 64)] | 295,168 |
| BatchNormalization | [(None, None, 241, 480, 64)] | 256 |
| ConvLSTM2D | [(None, None, 241, 480, 64)] | 33,024 |
| Conv3D | [(None, None, 241, 480, 1)] | 1,729 |

Total parameters: 753,089

Trainable parameters: 752,833

Non-trainable parameters: 256

Similarly, the model training process can be mathematically represented as an optimization problem:

$$\min_{\boldsymbol{\theta}_g} L\left(g(\tilde{\boldsymbol{\varphi}}_{t+\delta t}^f, \mathbf{y}_{t+\delta t} - \tilde{\boldsymbol{\varphi}}_{t+\delta t}^f, \boldsymbol{\theta}_g), \boldsymbol{\varepsilon}_{t+\delta t}^f\right), \quad t \in [T_1, T_2) \tag{4}$$

where $g$ represents the DANet, a nonlinear DA model, with the PredNet forecast $\tilde{\boldsymbol{\varphi}}_{t+\delta t}^f$ and its discrepancy from observation $\mathbf{y}_{t+\delta t} - \tilde{\boldsymbol{\varphi}}_{t+\delta t}^f$ at time $t + \delta t$ as inputs. $\boldsymbol{\theta}_g$ represents the trainable parameters of DANet. $\boldsymbol{\varepsilon}_{t+\delta t}^f$ represents the discrepancy between the PredNet forecast and the "ground truth", serving as the label in the training process. The MSE loss and the Adam optimization algorithm are used in the DANet training process. All samples for DANet training, including PredNet forecasts and observations as input and corresponding "ground truth" as output, were collected between $T_1$ and $T_2$.

Upon completing the training process, the DANet can be used to estimate PredNet forecast errors from PredNet forecasts and available observations:

$$\tilde{\boldsymbol{\varepsilon}}_{t+\delta t}^f = g(\tilde{\boldsymbol{\varphi}}_{t+\delta t}^f, \mathbf{y}_{t+\delta t} - \tilde{\boldsymbol{\varphi}}_{t+\delta t}^f, \boldsymbol{\theta}_g), \quad t > T_2, \tag{5}$$

where $\tilde{\boldsymbol{\varepsilon}}_{t+\delta t}^f$ represents the estimated PredNet forecast error at time $t + \delta t$. This estimated forecast error is further used to update the PredNet forecast to get an improved estimate for the system state:

$$\tilde{\boldsymbol{\varphi}}_{t+\delta t}^a = \tilde{\boldsymbol{\varphi}}_{t+\delta t}^f + \tilde{\boldsymbol{\varepsilon}}_{t+\delta t}^f, \tag{6}$$

where $\tilde{\boldsymbol{\varphi}}_{t+\delta t}^a$ is the improved estimate of the state variable, called "analysis state variable", at time $t + \delta t$. Operational forecasting involves iteratively conducting forecasts using the trained

PredNet and periodically performing DA using the trained DANet. Here, operational forecasting is conducted after $T_2$ to preserve the causality inherent in forecasting problems.

## 2.3 Implementation details

The proposed method consists of three key steps for implementation. Initially, we train the PredNet on historical data and make forecasts. Subsequently, we train the DANet using the PredNet forecasts and observations, and "ground truth". Finally, operational forecasting is conducted through an iterative process, utilizing the trained PredNet and DANet. Detailed implementation steps for the proposed D-DNet operational forecasting method can be found in the following pseudocode:

---

*Pseudocode:*

1. **Initial condition**: analysis state variable $\tilde{\boldsymbol{\varphi}}_{t_0}^a$ at time $t_0$.

2. **PredNet forecast**: iteratively forecast state variables $\tilde{\boldsymbol{\varphi}}_{t+\delta t}^f$ from $\tilde{\boldsymbol{\varphi}}_t^f$ using PredNet (Equation 2).

3. **Observation**: Collect and pre-process observations and remove outliers $\mathbf{o}_{t+\delta t}$ at time $t+k\delta t$, where $k$ is a constant, $k\delta t$ represents the selected DA frequency.

4. **DANet assimilation**: estimate the PredNet forecast error $\tilde{\varepsilon}_{t+k\delta t}^f$ at time $t+k\delta t$ by assimilating the new coming observations $\mathbf{o}_{t+\delta t}$ into the PredNet forecast $\tilde{\boldsymbol{\varphi}}_{t+k\delta t}^f$ using DANet (Equation 5); obtain the analysis state variable $\tilde{\boldsymbol{\varphi}}_{t+k\delta t}^a$ using Equation 6.

5. **Reinitialize**: use the estimated analysis state variable $\tilde{\boldsymbol{\varphi}}_{t+k\delta t}^a$ as the initial condition of the subsequent forecast.

6. **Operational forecasting**: iteratively conduct Steps 2 – 4 to achieve long-term operational forecasting.

---

## 4. Evaluation metrics

### Root Mean Square Error (RMSE):

RMSE is a standard way to measure the error of a model in predicting quantitative data. The formula for calculating RMSE is:

$$RMSE = \sqrt{\frac{1}{N}\sum_{i=1}^{i=N}\left(\varphi_i - \tilde{\varphi}_i\right)^2} \,, \tag{7}$$

where $\varphi_i$ and $\tilde{\varphi}_i$ represent the "ground truth" and predicted values, respectively; $N$ is the total number of pixels or instances being evaluated.

### Correlation coefficient (R):

The correlation coefficient, particularly the Pearson correlation coefficient, is used as a metric in regression tasks and can be calculated as follows:

$$R = \frac{\sum_{i=1}^{i=N} \left( \varphi_i - \overline{\varphi} \right) \left( \varphi_i - \overline{\tilde{\varphi}} \right)}{\sqrt{\sum_{i=1}^{i=N} \left( \varphi_i - \overline{\varphi} \right)^2 \left( \varphi_i - \overline{\tilde{\varphi}} \right)^2}} , \tag{8}$$

where $\overline{\varphi}$ and $\overline{\tilde{\varphi}}$ are the means of all the $\varphi_i$ and $\tilde{\varphi}_i$ values being evaluated.

### _Cumulative Accuracy Profile (CAP):_

Mathematically, let $\left\{ RMSE_{\min}, ..., RMSE_{\max} \right\}$ denotes the full range of RMSE values observed in all instances of the model forecasts, and $\left\{ R_{\min}, ..., R_{\max} \right\}$ denotes the full range of R values. For each RMSE value $RMSE_i$ within this range, we obtain the CAP by calculating the percentage of forecast instances where their RMSE values are less than or equal to $RMSE_i$:

$$P_{RMSE_i} = \left( \frac{1}{K} \sum_{k=1}^{k=K} 1_{\{RMSE_k < RMSE_i\}} \right) \times 100\% . \tag{9}$$

Similarly, for each R value $R_i$ within this range, we calculate the percentage of forecast instances where their R values are less than or equal to $R_i$:

$$P_{R_i} = \left( \frac{1}{K} \sum_{k=1}^{k=K} 1_{\{R_k > R_i\}} \right) \times 100\% , \tag{10}$$

where $K$ is the total number of forecast instances; $RMSE_k$ and $R_k$ are the RMSE and R values for the kth forecast instance, respectively. $1_{\{\cdot\}}$ is an indicator function that equals 1 when its condition is satisfied and 0 otherwise.

**Supplementary Text**

*5. 5-day PM2.5 and AOD550 forecasts: PreNet vs. CAMS atmospheric models*

In this study, we conducted a comparative analysis of the PredNet, an advanced machine learning approach, against the renowned physics-based CAMS model in forecasting PM2.5 concentrations and AOD550 values over 5 days. To ensure a fair comparison, we generated 5-day ahead PM2.5 and AOD550 forecasts at 00 and 12 UTC daily, employing identical initial AOD550 values as those within the CAMS global atmospheric composition forecasts dataset. Performance was assessed using two key metrics, Root Mean Square Error (RMSE) and the correlation coefficient (R), with EAC4 reanalysis data serving as the "ground truth". To avoid potential comparison bias caused by subjective preferences, we randomly selected 100 starting times over the entire year of 2019 for conducting 5-day forecasts. Fig. S3 presents the mean and standard deviation of RMSE and R of these forecasts. For both PM2.5 and AOD550, the PredNet exhibited superiority over the CAMS in both RMSE and R and consistently outperformed the CAMS over 5 days, indicating its effectiveness in accurately capturing PM2.5 and AOD550 behavior.

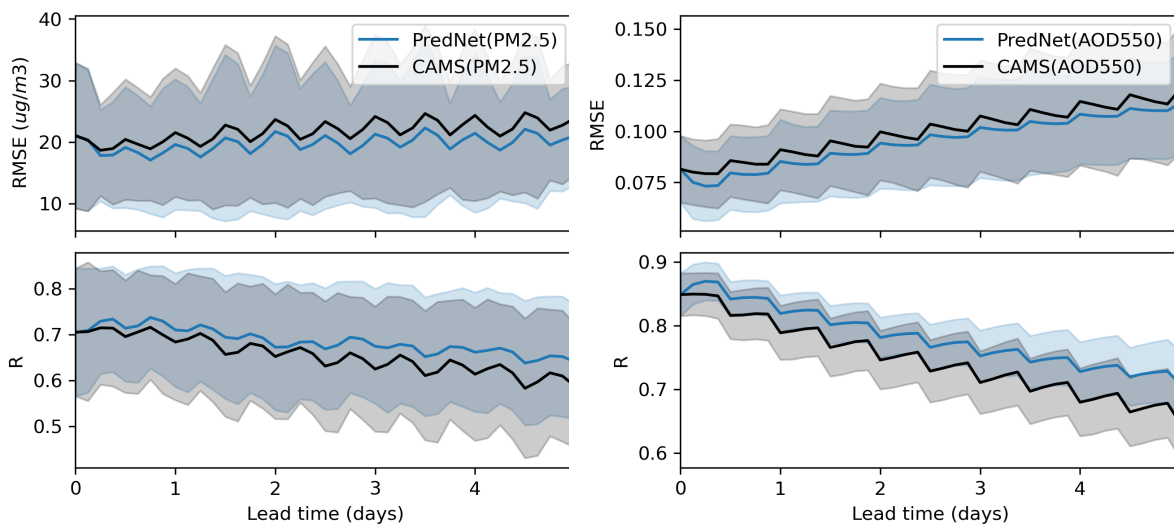

**Fig. S3. Comparison of forecasting performance** between the trained PredNet and the CAMS for PM2.5 concentrations and AOD550 values over a 5-day head time.

For further comparison, we display the PM2.5 concentrations and AOD550 values from EAC4 reanalysis data, PredNet forecasts, and CAMS forecasts at 1-day and 5-day lead times (Fig. S4-S5). This comparison indicates that both the PredNet and CAMS atmospheric models effectively capture the spatial distribution of PM2.5 concentrations and AOD550 values. Notably, high PM2.5 concentrations and AOD550 values are particularly observed over Asia in both the reanalysis and forecast datasets. PredNet consistently outperforms CAMS in forecasting both PM2.5 and AOD550 for the 1-day and 5-day periods. However, there is a clear trend that forecasting accuracy decreases as the forecast horizon extends from 1-day to 5-day ahead for both models. The RMSE and R values suggest that while both models can capture the general trend of the variations, there are limitations to their predictive abilities, especially as the forecast horizon increases.

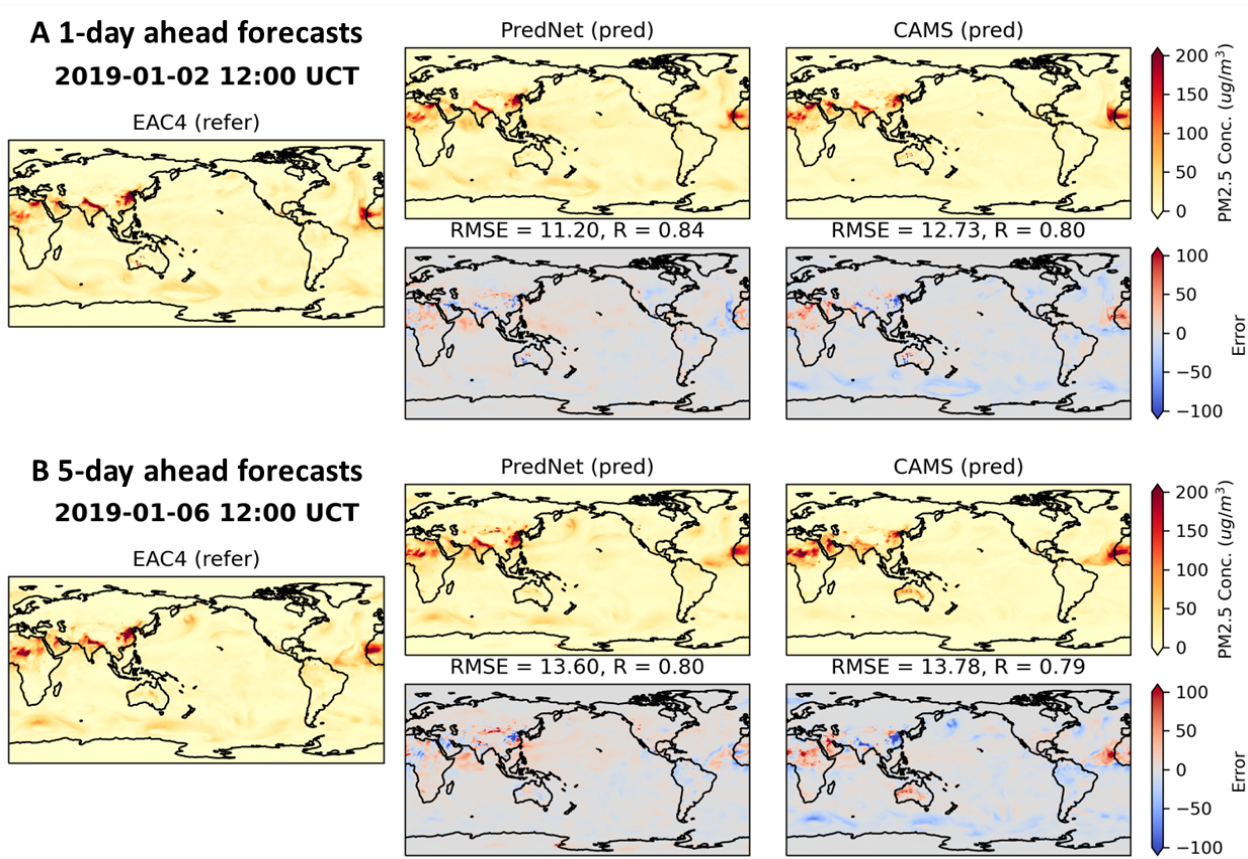

**Fig. S4. Comparison of PM2.5 forecasts** from PredNet and CAMS with the EAC4 references for (**A**) 1-day and (**B**) 5-day lead times. The forecasts are initialized from 2019-01-01 at 12:00 UTC.

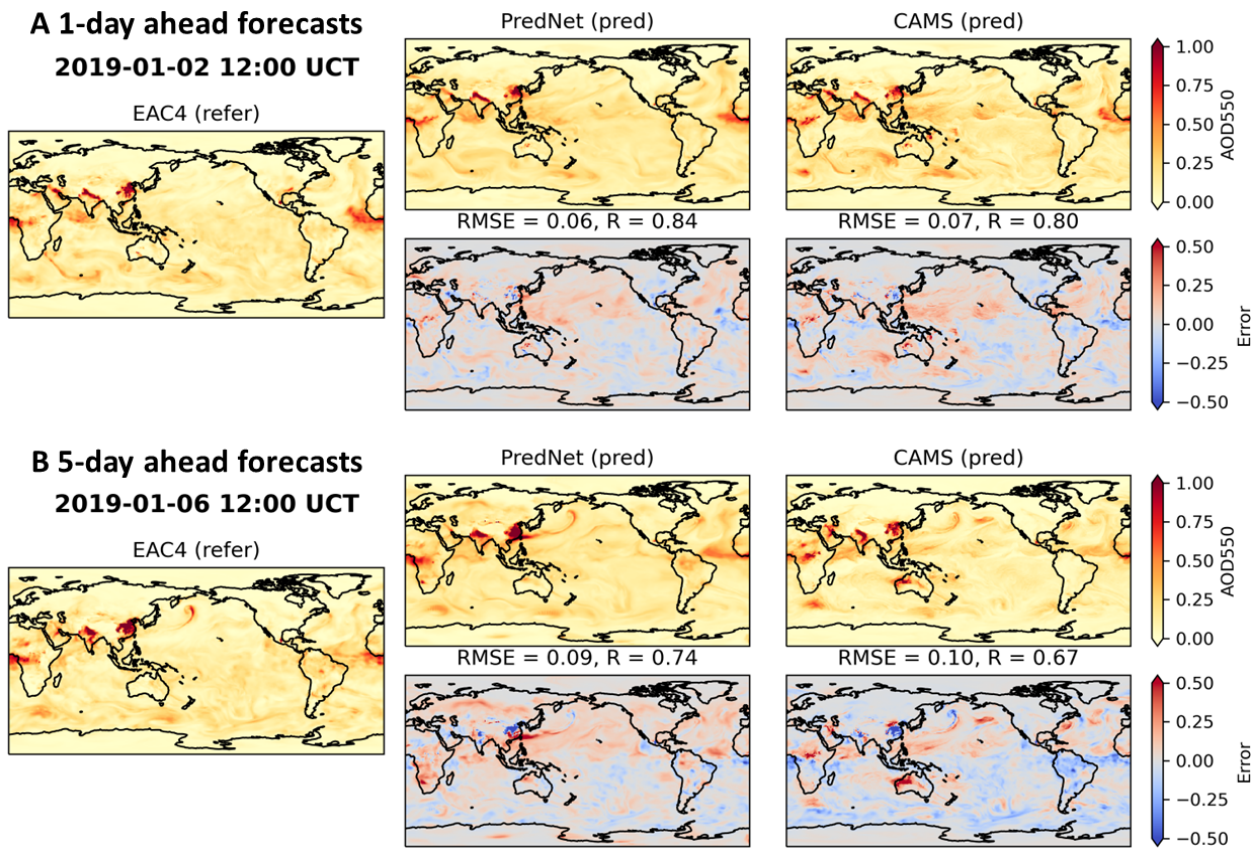

**Fig. S5. Comparison of AOD550 forecasts** from PredNet and CAMS with the EAC4 references for (**A**) 1-day and (**B**) 5-day lead times. The forecasts are initialized from 2019-01-01 at 12:00 UTC.

## *6. DANet and performance evaluation*

In this study, we employ Neural Network-based Data Assimilation (DANet) to enhance the accuracy of AOD550 forecasts. Before conducting DA, we compared the "ground truth" AOD550 from the EAC4 reanalysis dataset and the PredNet forecast at a 2-day lead time (Fig. S6). Additionally, AOD550 satellite observations and corresponding errors are included. For AOD550, reanalysis, forecasts, and satellite observations reveal high concentrations in similar regions, with associated error maps indicating smaller discrepancies compared to forecast errors. The comparison suggests spatial patterns in model accuracy and potential areas for improvement in forecasting results.

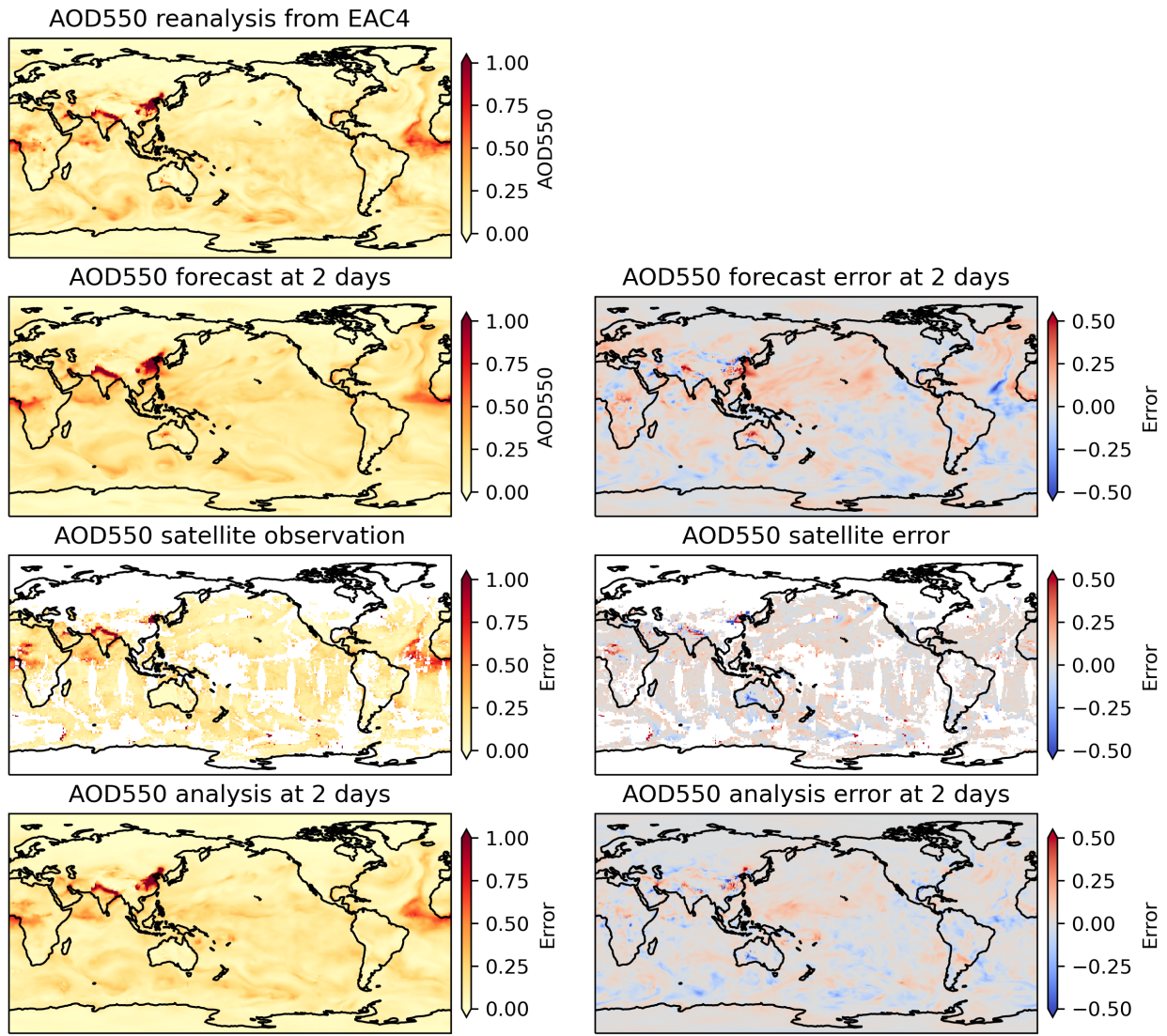

**Fig S6. Comparison of PredNet forecast, satellite observation, and DANet analysis for AOD550** at a 2-day lead time.

Through rigorous training, validation, and testing, the DANet effectively integrates satellite observations and PredNet forecasts, yielding reduced RMSE and improved R compared with the PredNet forecasts (Fig. S7). Analysis at a 2-day lead time revealed a notable reduction in error for AOD550 forecasts when compared to PredNet forecast errors (Fig. S6). These findings highlight the effectiveness of the DANet-based data assimilation process in enhancing forecast accuracy and reducing discrepancies, particularly in previously problematic regions exhibiting substantial forecast errors.

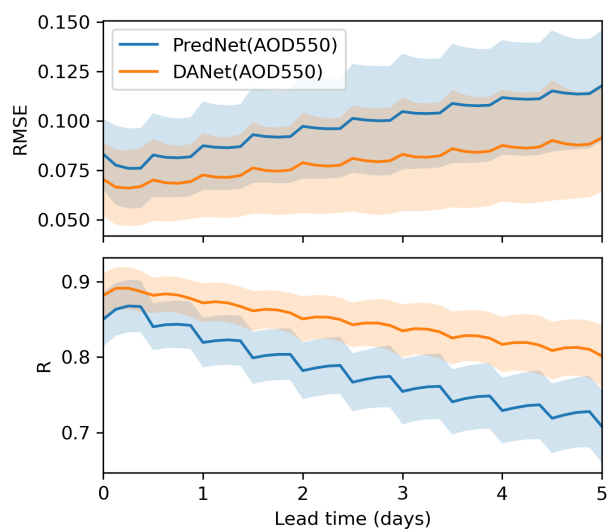

**Fig. S7. Comparison of AOD550 forecast and analysis results** before and after applying the proposed DANet.

## *7. Regional evaluation*

We performed a regional evaluation of the operational forecasting results generated by both D-DNet and the CAMS 4D-Var system. The regions and naming convention employed align with the ECMWF scorecards and GraphCast (Fig. S8) *(15)*. The per-region evaluation for RMSE and R metrics is displayed in Figures S9 to S12. The per-region CAP analysis for RMSE and R metrics is displayed in Figures S13 to S16. Our findings indicate that D-DNet delivers better forecasts compared with the CAMS 4D-Var system across the majority of regions over the entire year of 2019.

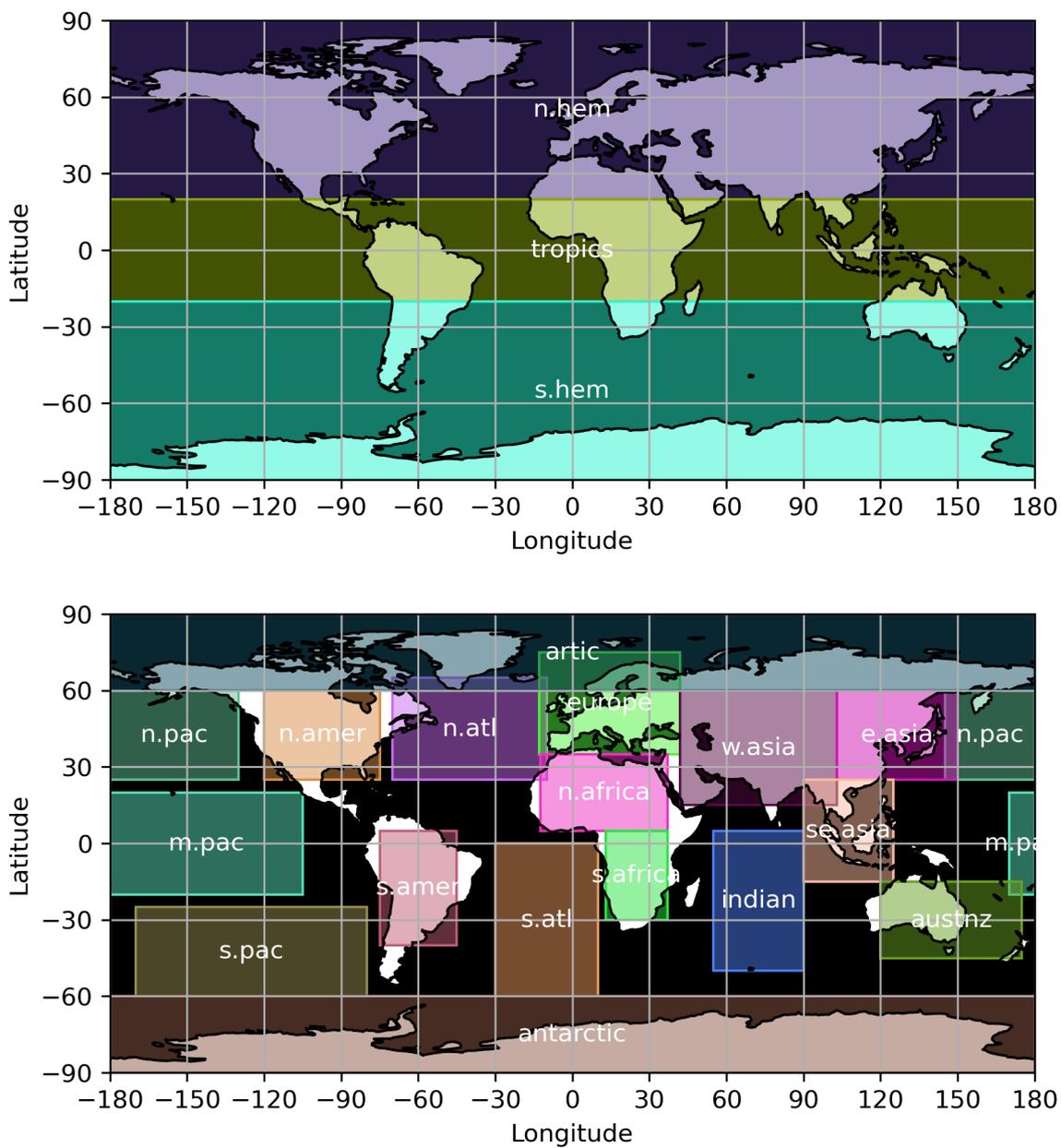

**Fig. S8: Regional specification for the regional analysis.** We use the same regions and naming convention as in the ECMWF scorecards https://sites.ecmwf.int/ifs/scorecards/

scorecards-47r3HRES.html) and in Lam (2023) *(15)*. Per-region evaluation for PM2.5 concentration and AOD550 is provided in Figures S10-S13.

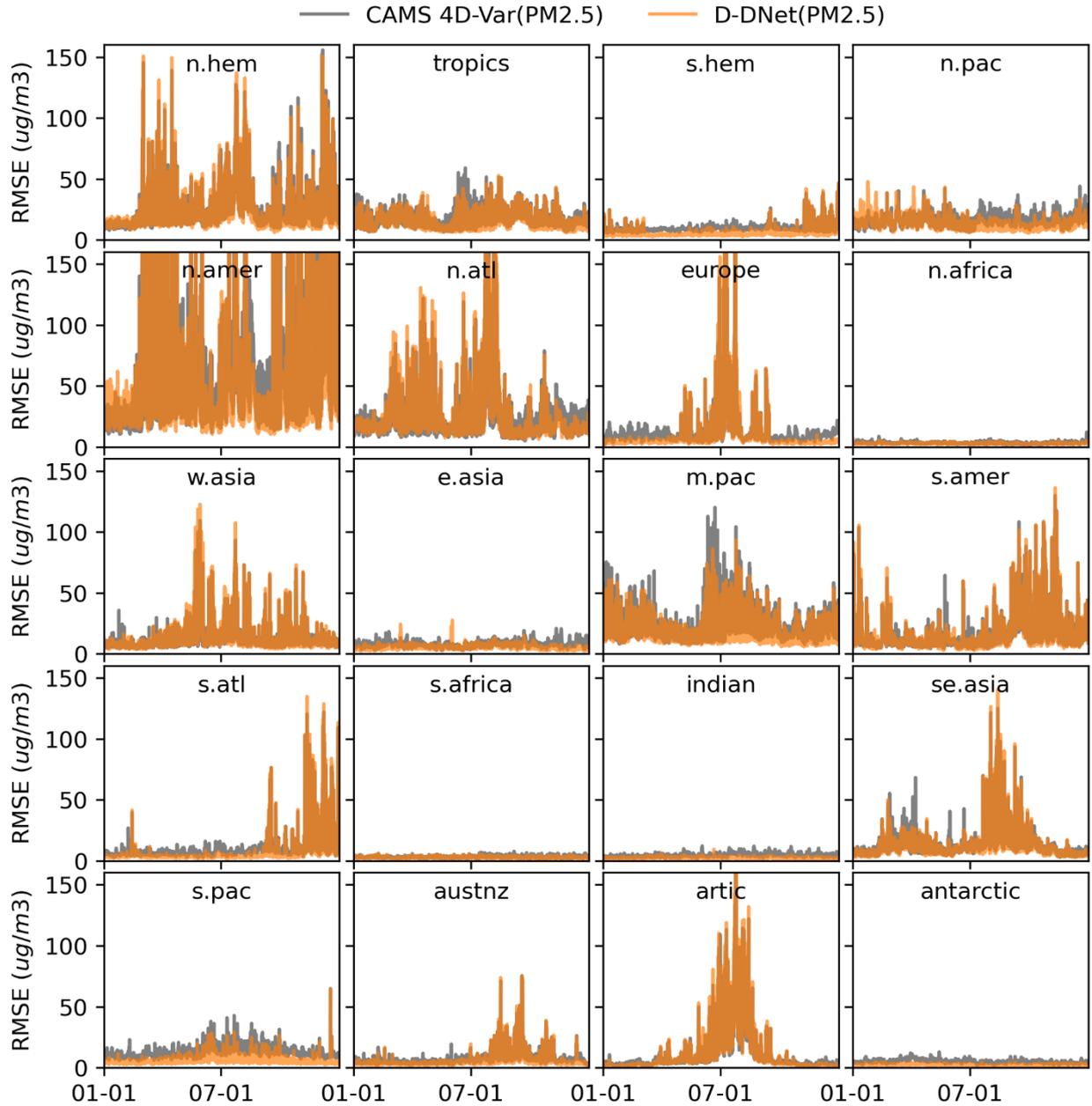

**Fig. S9. Regional analysis:** RMSE between PM2.5 concentration forecasts and "ground truth" in 2019 at different subregions. The operational forecasts from D-DNet are compared with those from the CAMS 4D-Var system, the renowned system for global atmospheric composition forecasting. (The subregions used here are displayed in Fig. S8.)

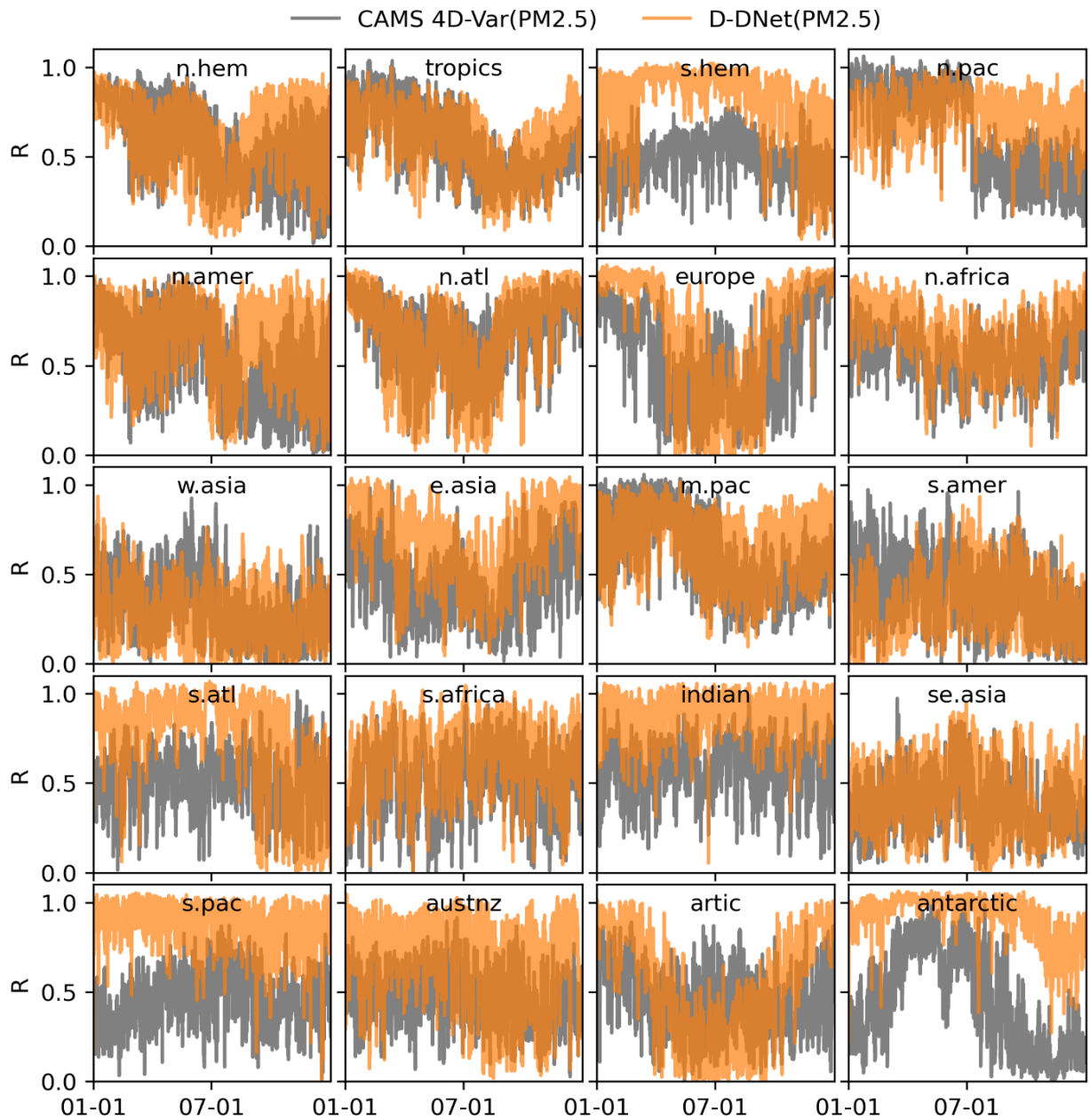

**Fig. S10.** The same as Fig. S9, but with regional R for PM2.5 forecasting displayed.

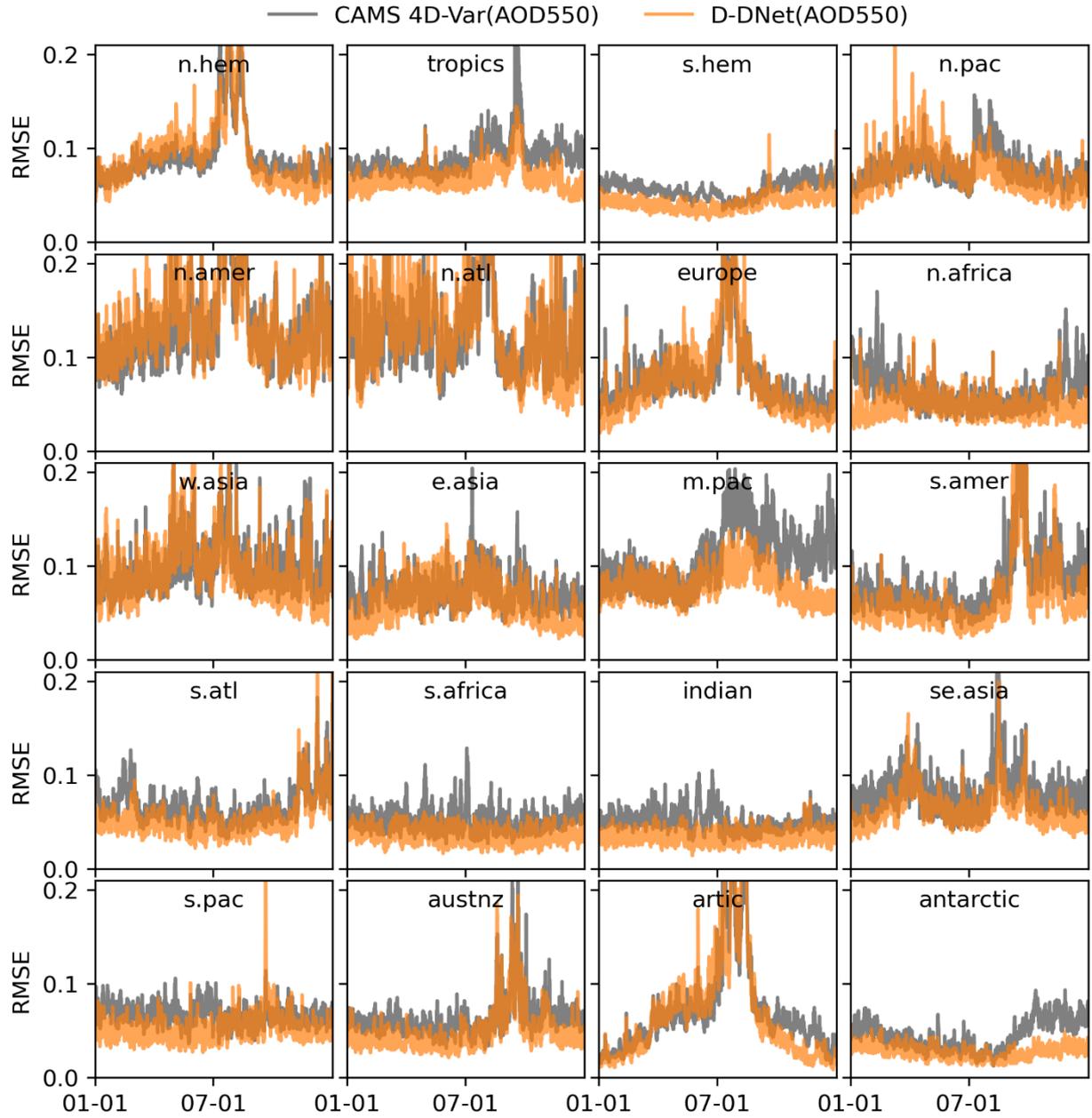

**Fig. S11.** The same as Fig. S9, but with regional RMSE for AOD550 forecasting displayed.

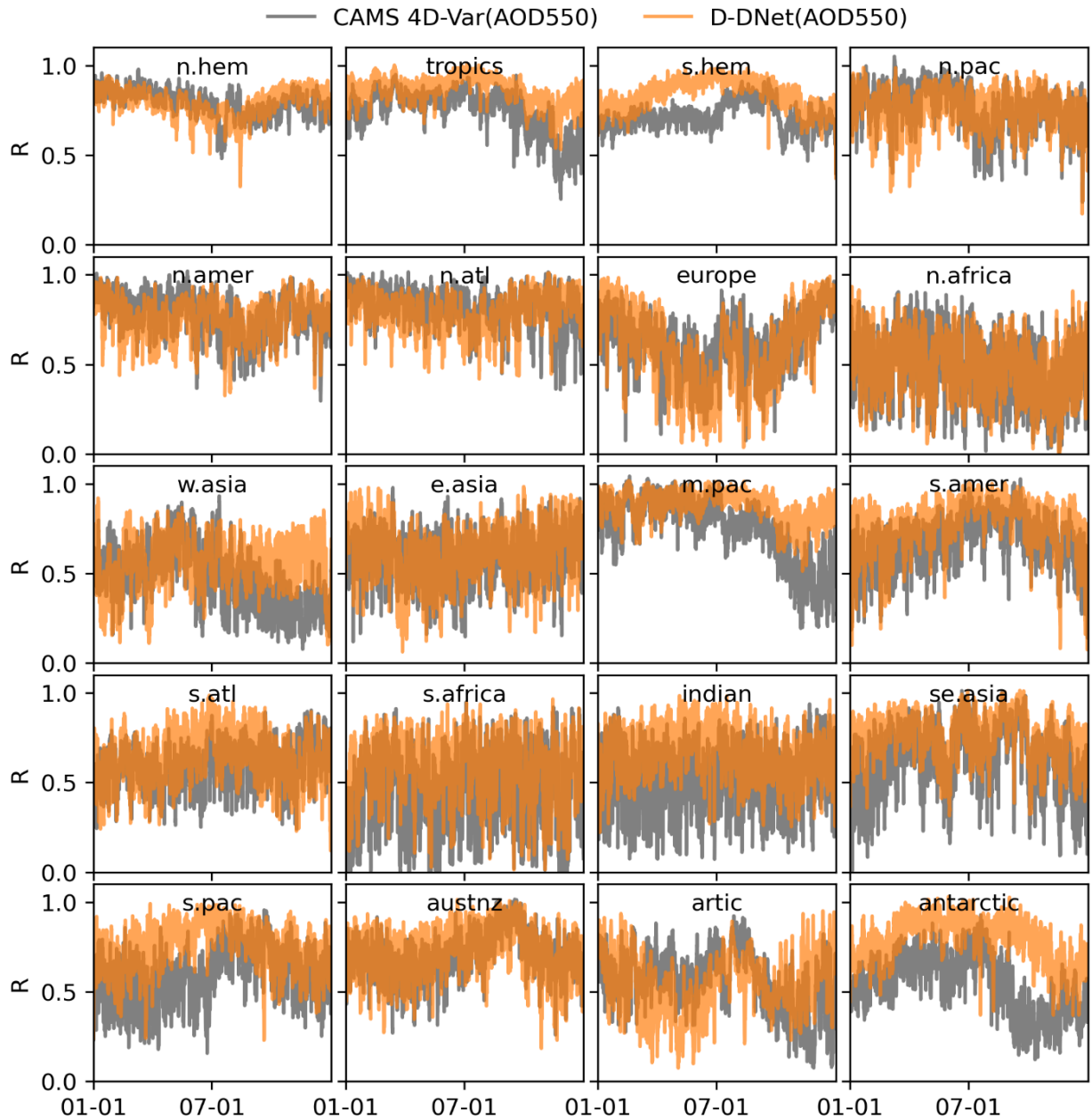

**Fig. S12.** The same as Fig. S9, but with regional R for AOD550 forecasting displayed.

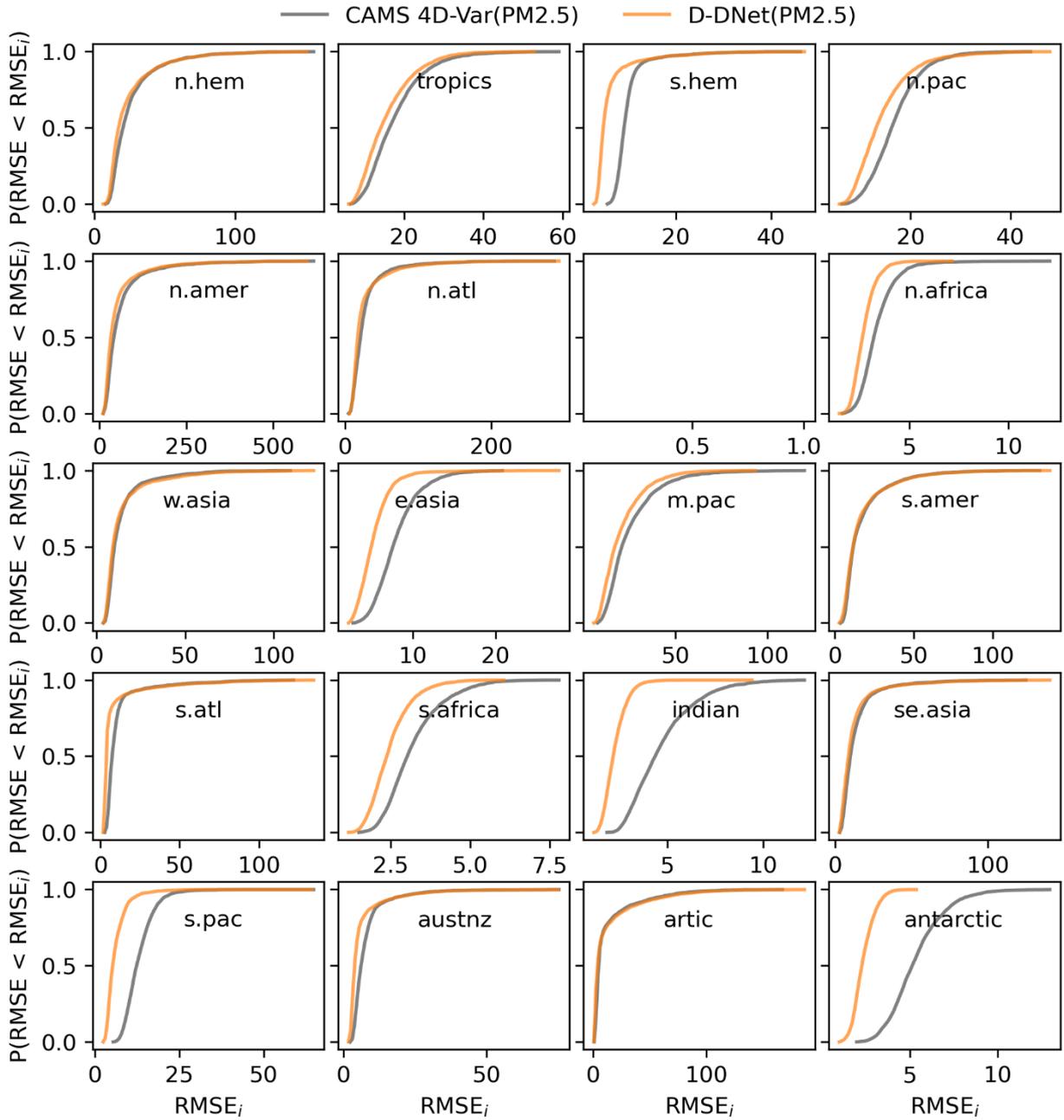

**Fig. S13. Cumulative Accuracy Profile (CAP) analysis** of D-DNet and the CAMS 4D-Var system for operational PM2.5 forecasting with respect to RMSE in 2019 at different subregions.

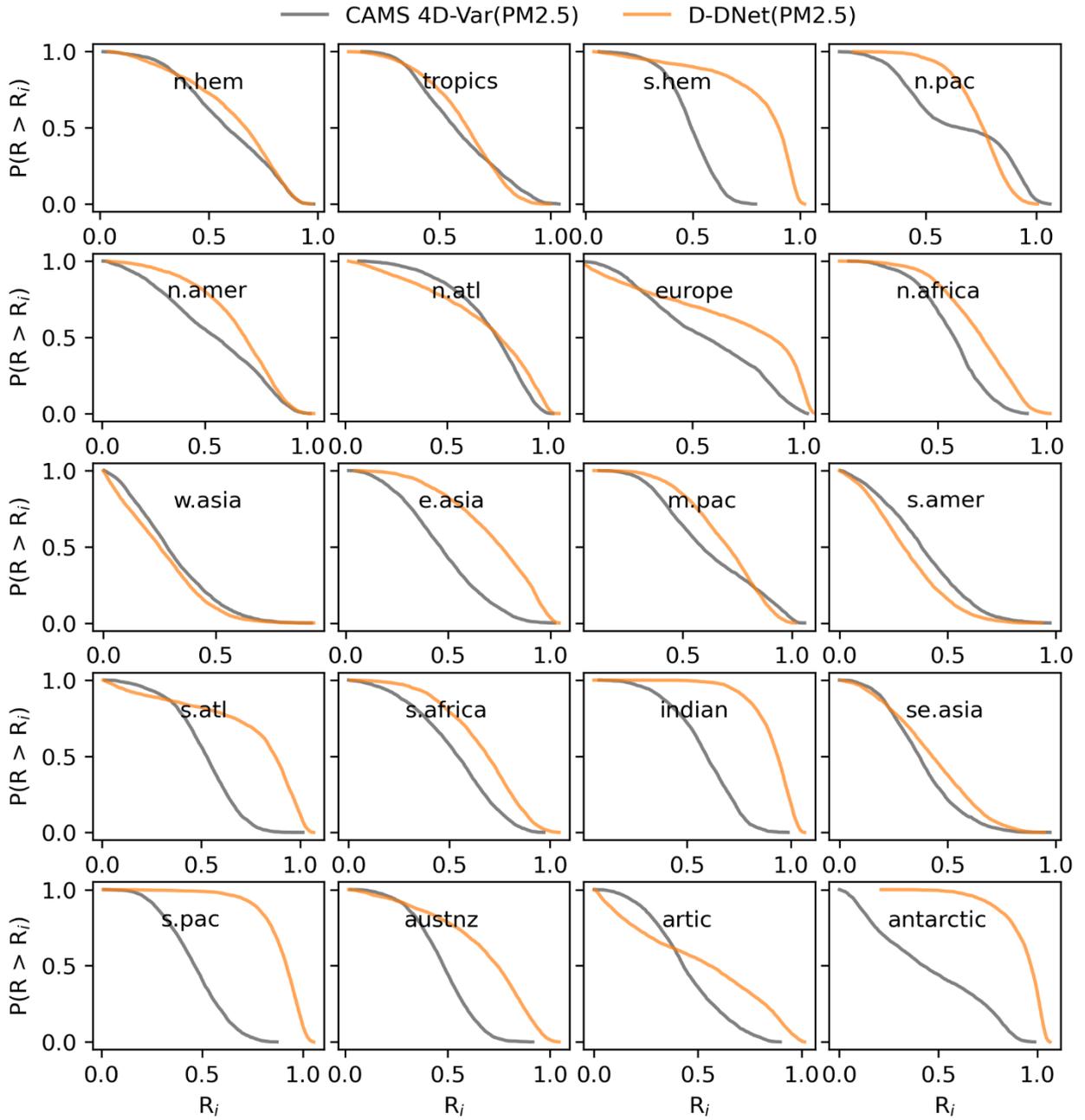

**Fig. S14. Cumulative Accuracy Profile (CAP) analysis** of D-DNet and the CAMS 4D-Var system for operational PM2.5 forecasting with respect to R in 2019 at different subregions.

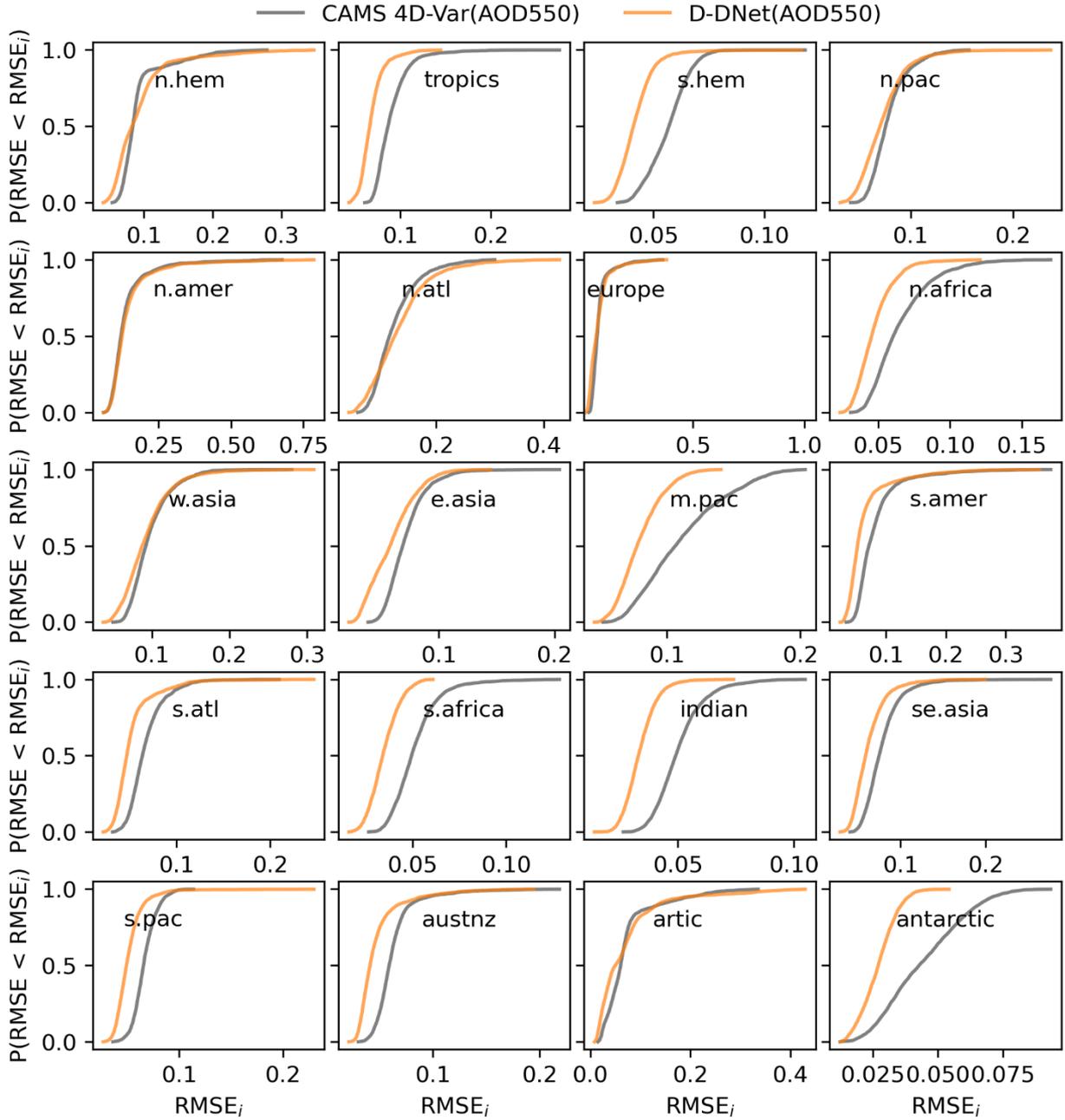

**Fig. S15. Cumulative Accuracy Profile (CAP) analysis** of D-DNet and the CAMS 4D-Var system for operational AOD550 forecasting with respect to RMSE in 2019 at different subregions.

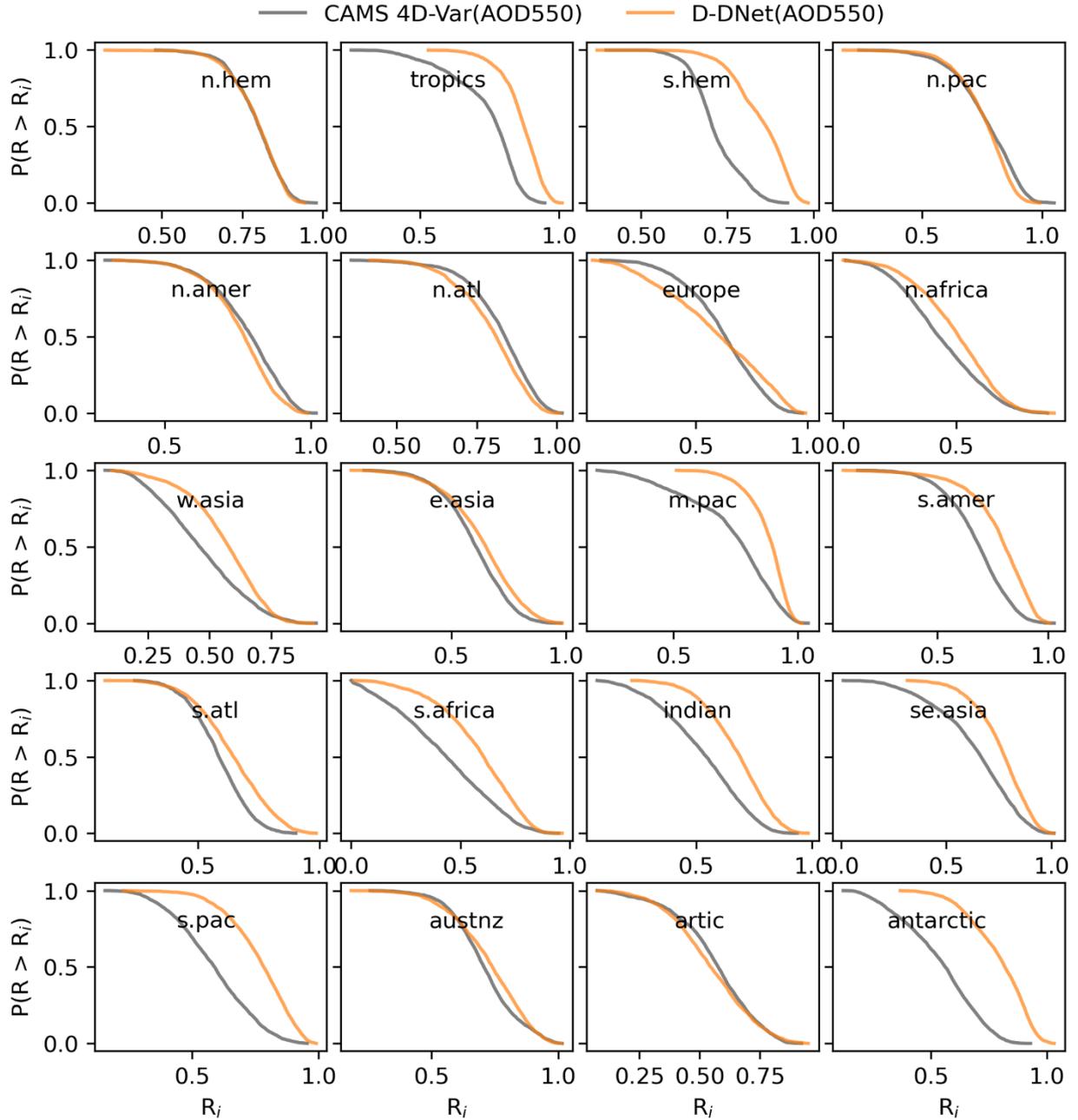

**Fig. S16. Cumulative Accuracy Profile (CAP) analysis** of D-DNet and the CAMS 4D Var system for operational AOD550 forecasting with respect to R in 2019 at different subregions.

## 8. Comparison between D-DNet and a neural network baseline (PredNet without DA)

Operational forecasting for PM2.5 and AOD550 employs D-DNet, comprising one network for initial forecasting (PredNet) and another for periodical data assimilation (DANet). For data assimilation, DANet incorporates real-time satellite observations as corrective data to refine the model outputs. By initializing the forecast model with updated PM2.5 and AOD550 forecasts derived from DA, we anticipate improving the accuracy of subsequent PM2.5 and AOD550 forecasts. This dual-network system ensures that operational forecasts remain up-to-date and reliable over a long period.

Figure S17 presents a comparative assessment of two forecasting models: D-DNet and PredNet, for predicting PM2.5 and AOD550 under identical initial conditions. As we extend the forecast time, we observe that PredNet forecasts accumulate errors and experience a decrease in accuracy. In contrast, the D-DNet model, which periodically incorporates satellite observations, demonstrates improved performance with smaller RMSE and larger R values. This enhancement in D-DNet performance can be attributed to its advanced integration of real-time satellite data, which likely improves its accuracy and prevents error divergence over time. This evidence highlights the effectiveness of the D-DNet in operational forecasting for PM2.5 and AOD550.

Notably, when relying solely on PredNet, the forecasts initially yield satisfactory results in the early hours. However, as time progresses, deviations from the "ground truth" occur due to the accumulation of forecast errors. This highlights the critical importance of DA and the need for correcting initial conditions throughout the forecasting process. By integrating observations and updating the forecasting results every 12 hours, our proposed D-DNet operational forecasting model provides stable and satisfactory forecasts spanning the entire one-year forecasting period. The average RMSE and R for operational PM2.5 forecasting are 18.04 ug/m3 and 0.55, respectively. The average RMSE and R for operational AOD550 forecasting are 0.07 and 0.76, respectively. These comparisons demonstrate the effectiveness of our approach in enhancing forecast accuracy and mitigating error accumulation.

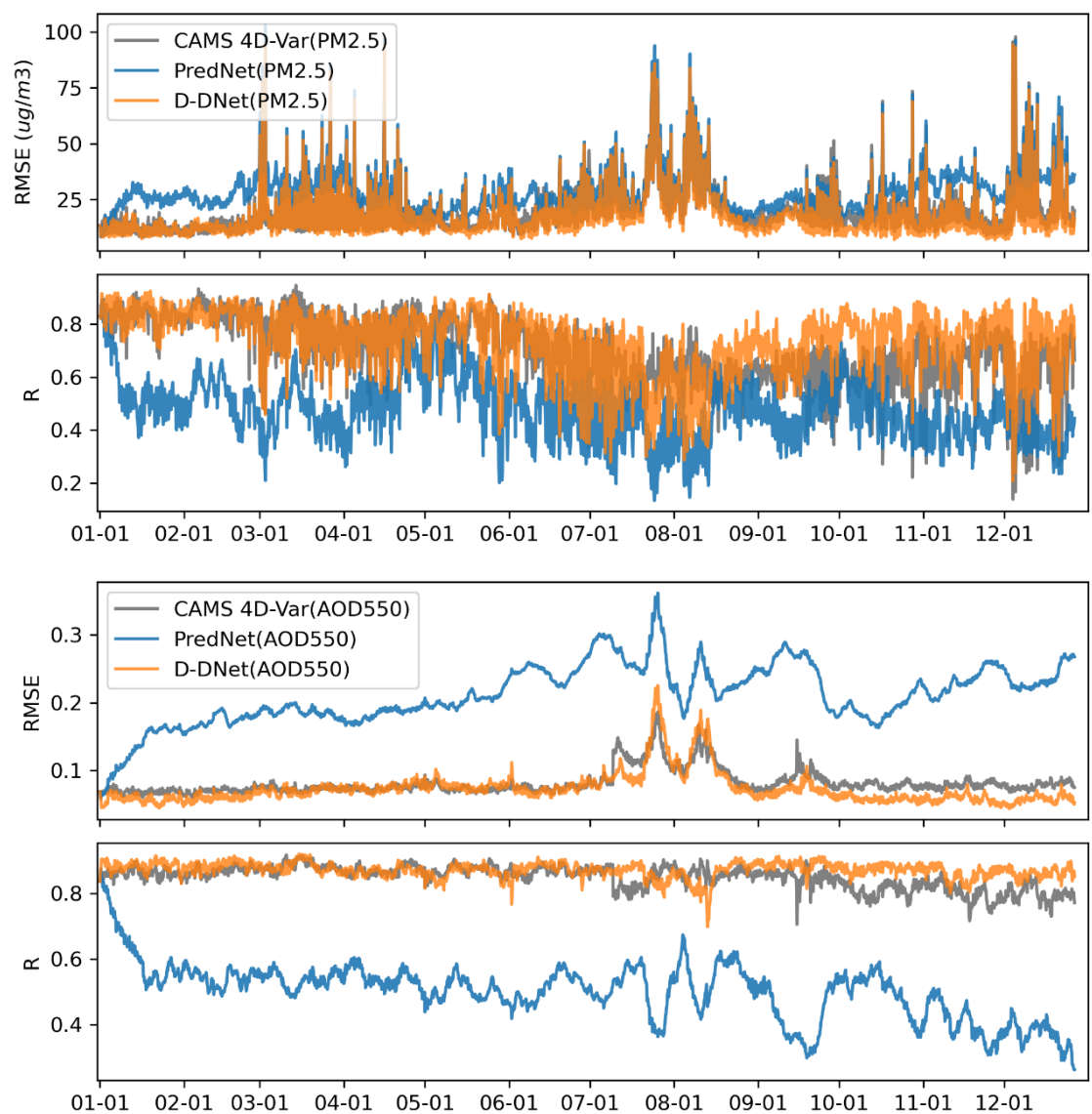

**Fig. S17. Comparison of operational PM2.5 and AOD550 forecasts** using D-DNet and PredNet, where PredNet serves as the baseline for a single neural network-based forecasting. The RMSE and R are computed against the EAC4 reanalysis dataset ("ground truth").

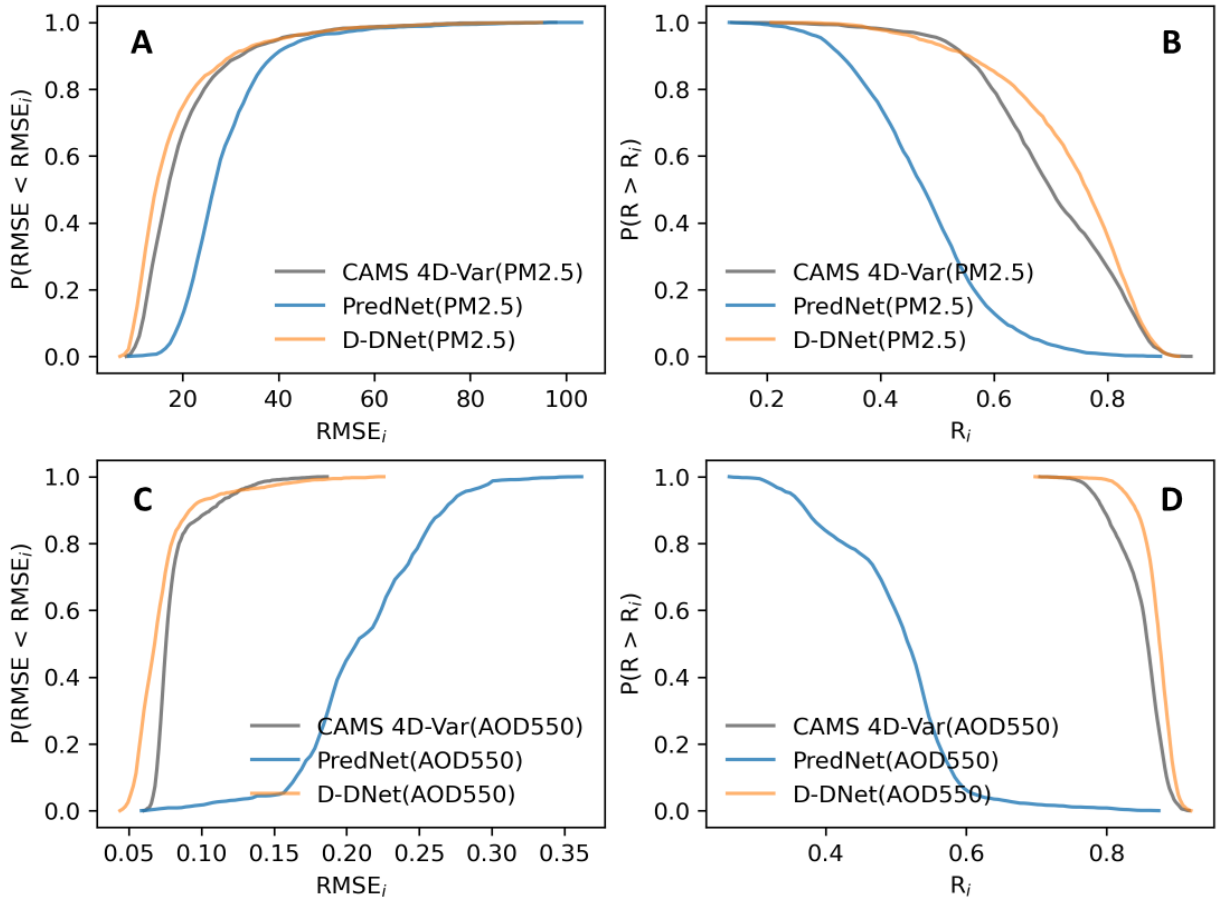

**Fig. 18. Cumulative Accuracy Profile (CAP) analysis** of D-DNet and PredNet for operational PM2.5 and AOD550 forecasting.

## 9. Computational efficiency

*D-DNet:* There are two Neural Networks contained in D-DNet: PredNet for forecasting and DANet for data assimilation. PredNet took about 20 hours, and DANet took about 3 hours to train on 2 NVIDIA A100 GPU (40G) devices. After D-DNet is trained, it takes minimal time for forecasting and DA. On a single NVIDIA A4000 GPU (16G) device, D-DNet can produce 5-day forecasting for PM2.5 and AOD550 with a spatial resolution of $0.75° \times 0.75°$ and a temporal resolution of 3-hour in under 40 seconds. For the whole operational workflow, including DA and 5-day forecasting, it only takes about 41 seconds.

*The CAMS 4D-Var system:* This system runs on ECMWF's ATOS HPC which is in Bologna, Italy. It consists of 7680 nodes (128 cores and 256GB of memory per node) and the CPU architecture is AMD Rome (2.5 GHZ). CAMS forecast runs on 32 nodes (using 32*128=4096 cores). The forecast resolution is TL511, with a time step of 900 seconds and 137 model levels. Since the update to the last IFS cycle 48r1 in June 2023, the 5-day operational global forecast itself has an average run time of 832 seconds (with a standard deviation of 40 seconds). Considering the whole workflow steps involved in the operational production of each 5-day forecast, which includes fetching observations, pre-screening observations, 4D-Var analysis, forecasting, and product dissemination, the total time would be around 2 hours.

## *10. Discussion*

In this study, we made most of our efforts to construct a fully data-driven operational forecasting framework based on our proposed D-DNet operational forecasting model. This work aims to tackle two challenges: 1. Current neural network-based forecasting models can produce highly accurate forecasts in the short term, but still suffer from error accumulation and accuracy loss with an extension of forecasting horizons. 2. Conventional ensemble or variational-based DA methods can improve forecasting accuracy by incorporating observations, but they are computationally intensive and time consuming. We innovatively adopt two neural networks for forecasting and DA. In this way, we can achieve efficient operational forecasting even in large-scale problems. Our proposed D-DNet demonstrated improved accuracy for operational PM2.5 and AOD550 forecasts on a global scale compared with the CAMS 4D-Var system, a renowned operational atmospheric component forecasting model. However, there are still several aspects that are expected to be improved in the future.

*Potential improvements by including other atmospheric components:* In this study, our consideration was limited to PM2.5 concentrations, AOD550 values and most relevant atmospheric variables due to constraints in computational resources. Enhancements in forecasting accuracy are anticipated through the incorporation of additional atmospheric elements, such as NO2, SO2, and others, owing to the chemical reactions occurring between these trace gases and aerosols.

*Potential improvements by 3D modding:* In this study, our focus was only on meteorological variables, atmospheric components, and emission sources at the surface. However, atmospheric is a 3D system, including horizontal and vertical transports and reactions. It is expected that considering vertical reactions by including meteorological variables and atmospheric components will further improve the forecasting accuracy.

*Potential improvements by adopting state-of-the-art neural networks:* Our proposed D-DNet consists of two neural networks, one for forecasting and the other for DA. Both neural networks are constructed using ConvLSTM layers. Although ConvLSTM produces satisfactory performance in our spatiotemporal forecasting and DA tasks, we still expect to further improve both performances by adopting more advanced neural networks, such as Graph Neural Networks (GNNs), and Generative Adversarial Networks (GANs).

**References**

1. Y. Xie, H. Dai, H. Dong, T. Hanaoka, T. Masui, Economic Impacts from PM 2.5 Pollution-Related Health Effects in China: A Provincial-Level Analysis. *Environ Sci Technol* **50**, 4836–4843 (2016).

2. X. Bu, Z. Xie, J. Liu, L. Wei, X. Wang, M. Chen, H. Ren, Global PM2.5-attributable health burden from 1990 to 2017: Estimates from the Global Burden of disease study 2017. *Environ Res* **197**, 111123 (2021).

3. P. Bauer, A. Thorpe, G. Brunet, The quiet revolution of numerical weather prediction. *Nature* **525**, 47–55 (2015).

4. C. J. Walcek, N. M. Aleksic, A simple but accurate mass conservative, peak-preserving, mixing ratio bounded advection algorithm with FORTRAN code. *Atmos Environ* **32**, 3863–3880 (1998).

5. Z. Wang, W. Sha, H. Ueda, Numerical modeling of pollutant transport and chemistry during a high-ozone event in northern Taiwan. *Tellus B: Chemical and Physical Meteorology* **52**, 1189–1205 (2000).

6. F. Liang, Q. Xiao, K. Huang, X. Yang, F. Liu, J. Li, X. Lu, Y. Liu, D. Gu, The 17-y spatiotemporal trend of PM 2.5 and its mortality burden in China. *Proceedings of the National Academy of Sciences* **117**, 25601–25608 (2020).

7. Q. Zhang, Y. Zheng, D. Tong, M. Shao, S. Wang, Y. Zhang, X. Xu, J. Wang, H. He, W. Liu, Y. Ding, Y. Lei, J. Li, Z. Wang, X. Zhang, Y. Wang, J. Cheng, Y. Liu, Q. Shi, L. Yan, G. Geng, C. Hong, M. Li, F. Liu, B. Zheng, J. Cao, A. Ding, J. Gao, Q. Fu, J. Huo, B. Liu, Z. Liu, F. Yang, K. He, J. Hao, Drivers of improved PM2.5 air quality in China from 2013 to 2017. *Proc Natl Acad Sci U S A* **116**, 24463–24469 (2019).

8. Z. Chen, D. Chen, C. Zhao, M. Kwan, J. Cai, Y. Zhuang, B. Zhao, X. Wang, B. Chen, J. Yang, R. Li, B. He, B. Gao, K. Wang, B. Xu, Influence of meteorological conditions on PM2.5 concentrations across China: A review of methodology and mechanism. *Environ Int* **139**, 105558 (2020).

9. S. Ravuri, K. Lenc, M. Willson, D. Kangin, R. Lam, P. Mirowski, M. Fitzsimons, M. Athanassiadou, S. Kashem, S. Madge, R. Prudden, A. Mandhane, A. Clark, A. Brock, K. Simonyan, R. Hadsell, N. Robinson, E. Clancy, A. Arribas, S. Mohamed, Skilful precipitation nowcasting using deep generative models of radar. *Nature* **597**, 672–677 (2021).

10. M. G. Jacox, M. A. Alexander, D. Amaya, E. Becker, S. J. Bograd, S. Brodie, E. L. Hazen, M. Pozo Buil, D. Tommasi, Global seasonal forecasts of marine heatwaves. *Nature* **604**, 486–490 (2022).

11. D. Kochkov, J. A. Smith, A. Alieva, Q. Wang, M. P. Brenner, S. Hoyer, Machine learning–accelerated computational fluid dynamics. *Proceedings of the National Academy of Sciences* **118** (2021).

12. M. Cheng, F. Fang, I. M. Navon, J. Zheng, X. Tang, J. Zhu, C. Pain, Spatio-Temporal Hourly and Daily Ozone Forecasting in China Using a Hybrid Machine Learning Model: Autoencoder and Generative Adversarial Networks. *J Adv Model Earth Syst* **14**, 1–26 (2022).

13. J. Pathak, S. Subramanian, P. Harrington, S. Raja, A. Chattopadhyay, M. Mardani, T. Kurth, D. Hall, Z. Li, K. Azizzadenesheli, P. Hassanzadeh, K. Kashinath, A. Anandkumar, FourCastNet: A Global Data-driven High-resolution Weather Model using Adaptive Fourier Neural Operators. arXiv:2202.11214v1 [physics.ao-ph] (2022).

14. K. Bi, L. Xie, H. Zhang, X. Chen, X. Gu, Q. Tian, Accurate medium-range global weather forecasting with 3D neural networks. *Nature* **619**, 533–538 (2023).

15. R. Lam, A. Sanchez-Gonzalez, M. Willson, P. Wirnsberger, M. Fortunato, F. Alet, S. Ravuri, T. Ewalds, Z. Eaton-Rosen, W. Hu, A. Merose, S. Hoyer, G. Holland, O. Vinyals, J. Stott, A. Pritzel, S. Mohamed, P. Battaglia, Learning skillful medium-range global weather forecasting. *Science (1979)* **382**, 1416–1421 (2023).

16. M. Reichstein, G. Camps-Valls, B. Stevens, M. Jung, J. Denzler, N. Carvalhais, Prabhat, Deep learning and process understanding for data-driven Earth system science. *Nature* **566**, 195–204 (2019).

17. H. Wu, H. Zhou, M. Long, J. Wang, Interpretable weather forecasting for worldwide stations with a unified deep model. *Nat Mach Intell* **5**, 602–611 (2023).

18. Y. Qi, Q. Li, H. Karimian, D. Liu, A hybrid model for spatiotemporal forecasting of PM 2.5 based on graph convolutional neural network and long short-term memory. *Science of the Total Environment* **664**, 1–10 (2019).

19. M. Leutbecher, Ensemble size: How suboptimal is less than infinity? *Quarterly Journal of the Royal Meteorological Society* **145**, 107–128 (2019).

20. A. Gettelman, A. J. Geer, R. M. Forbes, G. R. Carmichael, G. Feingold, D. J. Posselt, G. L. Stephens, S. C. van den Heever, A. C. Varble, P. Zuidema, The future of Earth system prediction: Advances in model-data fusion. *Sci Adv* **8**, 3488 (2022).

21. T. C. Vance, T. Huang, K. A. Butler, Big data in Earth science: Emerging practice and promise. *Science* **383** (2024).

22. A. Inness, M. Ades, A. Agustí-Panareda, J. Barré, A. Benedictow, A.-M. Blechschmidt, J. J. Dominguez, R. Engelen, H. Eskes, J. Flemming, V. Huijnen, L. Jones, Z. Kipling, S. Massart, M. Parrington, V.-H. Peuch, M. Razinger, S. Remy, M. Schulz, M. Suttie, The CAMS reanalysis of atmospheric composition. *Atmos Chem Phys* **19**, 3515–3556 (2019).

23. A. Soulie, C. Granier, S. Darras, N. Zilbermann, T. Doumbia, M. Guevara, J.-P. Jalkanen, S. Keita, C. Liousse, M. Crippa, D. Guizzardi, R. Hoesly, S. J. Smith, Global anthropogenic emissions (CAMS-GLOB-ANT) for the Copernicus Atmosphere Monitoring Service simulations of air quality forecasts and reanalyses. *Earth Syst Sci Data* **16**, 2261–2279 (2024).

24. C. Granier, S. Darras, H. D. van der Gon, J. Doubalova, N. Elguindi, B. Galle, M. Gauss, M. Guevara, J.-P. Jalkanen, J. Kuenen, C. Liousse, B. Quack, D. Simpson, K. Sindelarova, The Copernicus Atmosphere Monitoring Service global and regional emissions. *Copernicus Atmosphere Monitoring Service*, 54 (2019).

25. J. Handschuh, T. Erbertseder, M. Schaap, F. Baier, Estimating PM2.5 surface concentrations from AOD: A combination of SLSTR and MODIS. *Remote Sens Appl* **26**, 100716 (2022).

26.  Q. He, M. Wang, S. H. L. Yim, The spatiotemporal relationship between PM2.5 and aerosol optical depth in China: influencing factors and implications for satellite PM2.5 estimations using MAIAC aerosol optical depth. *Atmos Chem Phys* **21**, 18375–18391 (2021).

27.  R. C. Levy, L. A. Remer, D. Tanré, S. Mattoo, Y. J. Kaufman, "ALGORITHM FOR REMOTE SENSING OF TROPOSPHERIC AEROSOL OVER DARK TARGETS FROM MODIS: Collections 005 and 051: Revision 2; Feb 2009" (2009).

28.  M. Guevara, O. Jorba, C. Tena, H. Denier Van Der Gon, J. Kuenen, N. Elguindi, S. Darras, C. Granier, C. Pérez García-Pando, Copernicus Atmosphere Monitoring Service TEMPOral profiles (CAMS-TEMPO): Global and European emission temporal profile maps for atmospheric chemistry modelling. *Earth Syst Sci Data* **13**, 367–404 (2021).

29.  T. Warneke, A. Arola, A. Benedictow, Y. Bennouna, L. Blake, I. Bouarar, Q. Errera, H. J. Eskes, J. Griesfeller, L. Ilić, J. Kapsomenakis, M. Kouyate, B. Langerock, A. Mortier, M. Pitkänen, I. Pison, M. Ramonet, A. Richter, A. Schoenhardt, M. Schulz, J. Tarniewicz, V. Thouret, A. Tsikerdekis, C. Zerefos, "Validation report of the CAMS near-real-time global atmospheric composition service: September – November 2023" (2024); https://doi.org/10.24380/90z9-nva.

30.  S. Zhai, D. J. Jacob, J. F. Brewer, K. Li, J. M. Moch, J. Kim, S. Lee, H. Lim, H. C. Lee, S. K. Kuk, R. J. Park, J. I. Jeong, X. Wang, P. Liu, G. Luo, F. Yu, J. Meng, R. V. Martin, K. R. Travis, J. W. Hair, B. E. Anderson, J. E. Dibb, J. L. Jimenez, P. Campuzano-Jost, B. A. Nault, J. H. Woo, Y. Kim, Q. Zhang, H. Liao, Relating geostationary satellite measurements of aerosol optical depth (AOD) over East Asia to fine particulate matter (PM2.5): insights from the KORUS-AQ aircraft campaign and GEOS-Chem model simulations. *Atmos Chem Phys* **21**, 16775–16791 (2021).

31.  X. Shi, Z. Chen, H. Wang, D.-Y. Yeung, W. Wong, W. Woo, Convolutional LSTM Network: A Machine Learning Approach for Precipitation Nowcasting. *Adv Neural Inf Process Syst* **28** (2015).

32.  X. Shi, D.-Y. Yeung, Machine Learning for Spatiotemporal Sequence Forecasting: A Survey. arXiv:1808.06865v1 [cs.LG] (2018).

33.  M. Cheng, F. Fang, I. M. Navon, C. Pain, Ensemble Kalman filter for GAN-ConvLSTM based long lead-time forecasting. *J Comput Sci* **69**, 102024 (2023).

34.  C.-A. Diaconu, S. Saha, S. Gunnemann, X. Xiang Zhu, "Understanding the Role of Weather Data for Earth Surface Forecasting using a ConvLSTM-based Model" in *2022 IEEE/CVF Conference on Computer Vision and Pattern Recognition Workshops (CVPRW)* (IEEE, 2022; https://ieeexplore.ieee.org/document/9857012/)vols. 2022-June, pp. 1361–1370.

35.  G. Evensen, Sequential data assimilation with a nonlinear quasi-geostrophic model using Monte Carlo methods to forecast error statistics. *J Geophys Res* **99**, 10143–10162 (1994).

36.  A. Sandu, T. Chai, Chemical Data Assimilation—An Overview. *Atmosphere* **2**, 426–463 (2011).

37. S. K. Park, L. Xu, *Data Assimilation for Atmospheric, Oceanic and Hydrologic Applications (Vol. IV)* (Springer International Publishing, Cham, 2022; https://link.springer.com/10.1007/978-3-030-77722-7)vol. IV.

38. T. Miyoshi, K. Kondo, T. Imamura, The 10,240-member ensemble Kalman filtering with an intermediate AGCM. *Geophys Res Lett* **41**, 5264–5271 (2014).

39. A. C. Lorenc, The potential of the ensemble Kalman filter for NWP - A comparison with 4D-Var. *Quarterly Journal of the Royal Meteorological Society* **129**, 3183–3203 (2003).

40. G. Evensen, The Ensemble Kalman Filter: theoretical formulation and practical implementation. *Ocean Dyn* **53**, 343–367 (2003).